\documentclass[final]{cvpr}

\usepackage{times}
\usepackage{epsfig}
\usepackage{graphicx}
\usepackage{amsmath}
\usepackage{amssymb}
\usepackage{subfig}
\usepackage{enumitem}
\usepackage{bm}
\usepackage{color}
\usepackage{colortbl}
\usepackage{booktabs}
\definecolor{mygray}{gray}{.90}
\usepackage{appendix}
\usepackage{epstopdf}
\usepackage{multirow}
\usepackage{comment}
\usepackage{caption}
\usepackage[table,dvipsnames]{xcolor}
\usepackage{xspace}
\newcommand{\M}[1]{\mathbf{#1}}

\definecolor{wincolor}{rgb}{0.95, 0.2, 0.2}
\newcommand{\win}[1]{\textcolor{wincolor}{\bfseries{#1}}}
\newcommand{\second}[1]{\textcolor{NavyBlue}{\bfseries{#1}}}

\newcommand{\dataset}[1]{{\fontfamily{cmtt}\selectfont #1} }
\newcommand{\KITTI}{\dataset{KITTI}}
\newcommand{\Malaga}{\dataset{Malaga}}
\newcommand{\ETH}{\dataset{ETH3D}}

\def\gb{Gr{\"o}bner basis\xspace}

\usepackage[pagebackref=true,breaklinks=true,colorlinks,bookmarks=false]{hyperref}

\def\cvprPaperID{5280} 
\def\confYear{CVPR 2021}

\begin{document}

\title{Globally Optimal Relative Pose Estimation with Gravity Prior}

\author{Yaqing Ding$^{1}$, Daniel Barath$^{2,3}$, Jian Yang$^{1}$, Hui Kong$^{1}$, Zuzana Kukelova$^{2}$\\
$^1$ School of Computer Science and Engineering, 
Nanjing University of Science and Technology\\
$^2$ Visual Recognition Group, Faculty of Electrical Engineering, 
Czech Technical University in Prague \\
$^3$ Machine Perception Research Laboratory, 
SZTAKI, Budapest \\
{\tt\small dingyaqing@njust.edu.cn}
}

\maketitle

\begin{abstract}
	Smartphones, tablets and camera systems used, e.g., in cars and UAVs, are typically equipped with IMUs (inertial measurement units) that can measure the gravity vector accurately. 	
	Using this additional information, the $y$-axes of the cameras can be aligned, reducing their relative orientation to a single degree-of-freedom. With this assumption, we propose a novel globally optimal solver, minimizing the algebraic error in the least squares sense, to estimate the relative pose in the over-determined case. 
	Based on the epipolar constraint, we convert the optimization problem into solving two polynomials with only two unknowns.
	Also, a fast solver is proposed using the first-order approximation of the rotation. 
	The proposed solvers are compared with the state-of-the-art ones on four real-world datasets with approx.\ 50000 image pairs in total.
	Moreover, we collected a dataset, by a smartphone, consisting of 10933 image pairs, gravity directions and ground truth 3D reconstructions. 
\end{abstract}

\section{Introduction}\label{sec:intro}


Finding the relative pose between two cameras is one of the fundamental geometric vision problems with many applications, for example, in structure-from-motion~\cite{rother2003multi, pollefeys2004visual, pollefeys2008detailed, agarwal2011building, wu2013towards, schonberger2016structure}, visual localization~\cite{zeisl2015camera,svarm2016city}, and SLAM~\cite{mur2017orb}. The robust relative pose estimation is usually done in a hypothesize and test framework, such as RANSAC~\cite{fischler1981random}.

Locally optimized (LO) RANSAC~\cite{2003Locally,lebeda2012fixing} and its variants, such as Graph-Cut RANSAC~\cite{barath2017graph}, USAC~\cite{Raguram2013USAC}, have shown a significant improvement in terms of accuracy and convergence
compared to the standard RANSAC~\cite{fischler1981random}. 
The key idea of these RANSAC variants is to apply a local optimization step to refine each so-far-the-best model using non-minimal samples. The main benefit of using a non-minimal sample is that it allows for averaging out observational noise in the measurements. 
Besides improving the accuracy, this often speeds up the robust estimation via finding an accurate model early and, thus, triggering the termination criterion.
Both minimal and non-minimal solvers play important roles in the LO-RANSAC framework. 

Minimal solutions to the relative pose estimation with different camera configurations have been well-studied in the literature. For instance, given two calibrated cameras, the relative pose can be efficiently recovered from five point correspondences~\cite{nister2004efficient,kukelova2012polynomial,hartley2012efficient}. Recently, point plus direction-based methods have shown benefits. The basic idea of such methods is to align one axis (\eg, $y$-axis) of the cameras with the common reference direction, \eg, the gravity vector. The relative orientation reduces to 1-DoF, and the relative pose can be estimated from only three point correspondences~\cite{fraundorfer2010minimal,naroditsky2012two,sweeney2014solving,saurer2017homography,ding2019efficient,ding_2020icra}.

\begin{figure}[t]
    \centering
	    \includegraphics[width=0.99\columnwidth]{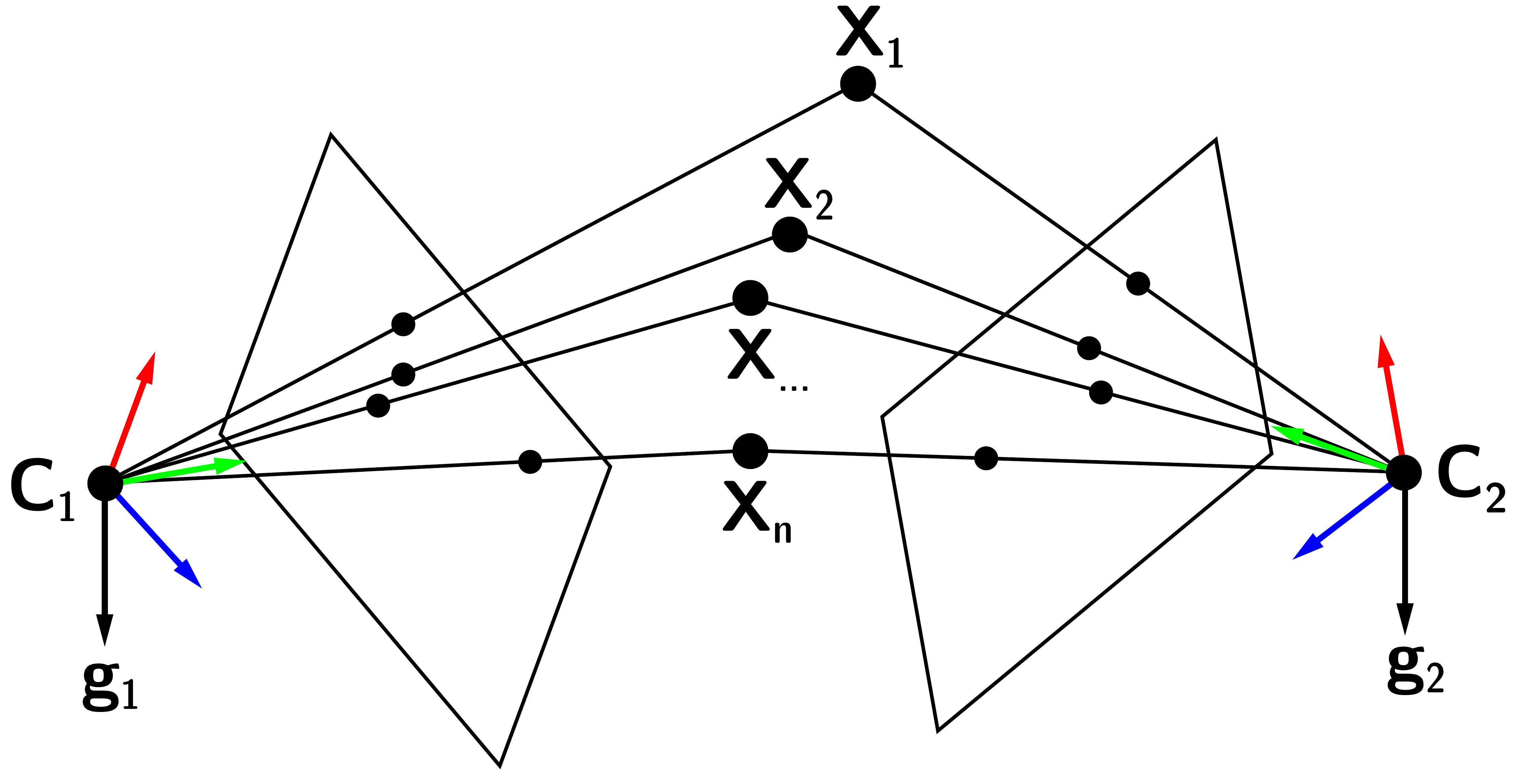}
        \caption{Two cameras $\M C_1$ and $\M C_2$ with known gravity directions $\M g_1$, $\M g_2$ observing $n \geq 4$ points $\M X_1, ..., \M X_n$. The objective is to find the relative pose of the cameras. }
    \label{fig:illustration}
\end{figure}


While the problem of estimating the absolute camera pose from a non-minimal set of correspondences (the PnP problem) has received a lot of attention and many efficient and globally optimal solutions were published \cite{hesch2011direct,zheng2013revisiting,kneip2014upnp,nakano2015globally,nakano2016versatile,wang2018simple,zhou2019efficient,terzakis2020consistently},
this does not hold for the relative pose estimation problem, also known as the \emph{N-point problem}. 
The main reason is the fact that the  N-point problem results in a more complicated optimization than the PnP problem.

The most commonly used non-minimal relative pose solver is  the
well-known direct linear transformation (DLT) method~\cite{hartley2003multiple}. This method can be used to estimate the essential or the fundamental matrix from 8 (the well-known 8-point algorithm~\cite{hartley1995defence}) or more point correspondence. The DLT method assumes that the matrix elements are linearly independent.  
Therefore, this method does not minimize a properly defined error in the 5D space of essential matrices for calibrated cameras, and a correct essential matrix has to be recovered from the approximate DLT solution~\cite{hartley2003multiple}.

Therefore, several approaches that are directly minimizing an energy function defined in the 
5D space of relative poses have been published recently. However, these approaches either do not guarantee a global optimum~\cite{hartley98iccv,helmke2007essential,ma2001optimization} or are too slow for practical applications~\cite{HartleyICCV07,chesi2008camera,hartley2009global}. 

Kneip et al. \cite{Kneip2013DirectOO} presented an eigenvalue-based formulation that is minimizing an algebraic error for the relative pose problem, which is after eliminating the relative translation, parameterized in the 3-dimensional space of relative rotations. However, the proposed iterative Levenberg-Marquardt solution does not guarantee optimality and it requires a reasonably good initialization.
The certifiably globally optimal solution to the eigenvalue-based formulation of relative pose estimation problem was proposed in~\cite{Briales_2018_CVPR}. 
This solution minimizes the algebraic error in the space of rotations and translations and formulates the problem as a quadratic program (QCQP) that is solved using a Semidefinite Program (SDP) relaxation. While the proposed method does find the global optimum of the cost function, it is very time-consuming with running time approximately $1s$. Moreover, this method does not provide good pose estimates for forward motion. The solution was later extended to globally optimal essential matrix  estimation for generalized cameras~\cite{Zhao_2020_CVPR}.
The globally optimal solution for the N-point problem~\cite{Briales_2018_CVPR} was recently improved in~\cite{zhao2020pami}.
The proposed formulation results in QCQP, however, in fewer variables and constraints than the formulation in~\cite{Briales_2018_CVPR}.
The final solver is $1$-$2$  orders of magnitude faster than~\cite{Briales_2018_CVPR}. 

In this paper, to estimate the relative pose globally optimally in real-time, we assume that the views share a common reference direction. This case is relevant since smartphones, tablets and camera systems used, \eg, in cars and UAVs, are typically equipped with IMUs (inertial measurement units) that can measure the gravity vector accurately. With this assumption, the relative rotation is reduced to 1-DoF and the N-point relative pose problem is significantly simplified.
The main contributions of this paper are:
\begin{itemize}
\vspace{-0.7em}
    \item 
    We propose a  {\bf novel real-time globally optimal solver} that minimizes the algebraic error in the least squares sense and estimates the relative pose of calibrated cameras with known gravity direction from N-point correspondences ($N\geq 4)$.
	Based on the epipolar constraint, we convert the optimization problem into solving two polynomials with only two unknowns and we propose two different efficient solutions to such a system.
	\vspace{-0.6em}
	\item In addition, we consider that for mobile robots, autonomous driving cars and UAVs, the relative rotation is small and, thus, we propose a {\bf solver estimating the linearized rotation efficiently}.
	\vspace{-0.6em}
	\item In extensive real and synthetic experiments we show an {\bf improvement in terms of accuracy and speed} over the state-of-the-art non-minimal N-point relative pose solvers. Moreover, we present a {\bf novel dataset} with 10993 image pairs taken by a smartphone with known gravity direction and ground truth 3D reconstructions. 
\end{itemize}

\section{Background}\label{section3}
Suppose that we are given 3D point $\M X_i$ observed in two calibrated views. Let $\M m_i=[u_i,v_i,1]^\top$ and $\M m'_i=[u'_i,v'_i,1]^\top$ be its projections in the two images in their homogeneous form. Since the gravity direction can be calculated from, {\eg} an IMU, we can, without loss of generality, align the $y$-axes of the cameras with the gravity direction. 
These alignments are done by rotation matrices $\M R$ and $\M R'$, giving rotated image points $\M p'_i = \M R' \M m'_i,\ \; \M p_i = \M R \M m_i$.  
After applying the rotations to the projected 2D points, we get 
\begin{equation}
\lambda' _i \M p'_i = \lambda _i \M R_y \M p_i +\M t , \label{q1}
\end{equation}
with 
%
%
\begin{equation}
\M R_y=\begin{bmatrix}
\cos \theta & 0 & \sin \theta\\
0 & 1 & 0\\
-\sin \theta & 0 & \cos \theta
\end{bmatrix},\label{q4}
\end{equation}
where $\theta$ is the unknown rotation angle around the vertical axis after the alignment by the gravity direction, $\M t$ is the unknown translational vector and $\lambda _i,\lambda'_i$ are unknown depths.
Vectors
$\lambda' _i \M p'_i$, $\lambda _i\M R_y \M p_i$ and $\M t$ are always coplanar. Therefore, the scalar triple product of these three vectors should be zero, and we obtain 
\begin{equation}
([\M p'_i]_\times \M R_y \M p_i)^\top \M t =0 . \label{q3}
\end{equation}
In this case, the depth parameters $\lambda_i, \lambda'_i$ are eliminated. 
%
%
Rotation matrix $\M R_y$~\eqref{q4} can be reparameterized as
\begin{equation}
\M R_y=\frac{1}{1+y^2}\begin{bmatrix}
1\!-\!y^2 & 0 & 2y\\
0 & 1\!+\!y^2 & 0\\
\!-\!2y & 0 & 1\!-\!y^2
\end{bmatrix},\label{q5}
\end{equation}
where $y = \tan \frac{\theta}{2}$. This parameterization introduces a degeneracy for a $180^\circ$ rotation which can be ignored in real applications~\cite{larsson2017efficient,Ding_2020_CVPR}.
The objective is to estimate the relative pose parameters $\theta$ and $\M t$ 
globally optimally in the over-determined case, {\ie}, 
from $N \geq 4$ point correspondences. 


\section{Problem Formulation}

Let us assume that we are given $N$ point correspondences.
By stacking the constraints~\eqref{q3}  for $N$ correspondences in a matrix, we obtain
\begin{equation}
\M A^\top \M t =0 , \label{q6}
\end{equation}
where $\M A$ is a $3\times N$ matrix with its $i^{th}$ column of form
\begin{equation}
\M A_i =[\M p'_i]_\times \M R_y \M p_i. \label{q7}
\end{equation}
The minimal case $N=3$ results in $\det(A)=0$ with up to 4 solutions, which has been well-studied in the literature~\cite{fraundorfer2010minimal,naroditsky2012two,ding_2020icra,sweeney2014solving}. We mainly focus on the case where $N\geq 4$. In the case when the point correspondences are contaminated by noise, the objective is to find the least squares optimal solution of~\eqref{q6}. 
The problem is formalized as follows:
\begin{equation}
\arg \limits_{\M R_y,\M t}\ \min \M t^{\top}\M C\ \M t, \label{q8}
\end{equation}
where $\M C=\M A \M A^\top$ is a $3\times3$ symmetric matrix
\begin{equation}
\M C=\begin{bmatrix}
c_{11} & c_{12} & c_{13}\\
c_{12} & c_{22} & c_{23}\\
c_{13} & c_{23} & c_{33}
\end{bmatrix}.\label{q8a}
\end{equation}
The elements of C are univariate polynomials in $y$. Eq.~\eqref{q8} is equivalent to minimizing the smallest eigenvalue of $\M C$ as
\begin{equation}
\arg \limits_{\M R_y}\ \min \alpha_{\min}(\M C). \label{q9}
\end{equation}
Note, that~\eqref{q9}  is a similar eigenvalue-based formulation to the one used in~\cite{Kneip2013DirectOO}. However, thanks to the special form of $\M R_y$, it results in a simpler optimization problem in 1D space instead of the 3D space of full rotations. Therefore, it can be solved globally optimally by efficiently computing all its stationary points, compared to~\cite{Kneip2013DirectOO} where an iterative solution was proposed.
The three eigenvalues of $3\times 3$ matrix $\M C$ should satisfy the following cubic constraint
\begin{equation}
\alpha^3-f_1\alpha^2+f_2\alpha-f_3=0, \label{q10}
\end{equation}
where
\begin{equation}
\begin{split}
f_1&={\operatorname {trace}}(\M C),\\
f_2&=c_{11}c_{22} + c_{11}c_{33} + c_{22}c_{33} - c_{12}^2-c_{13}^2-c_{23}^2, \\
f_3&=\det(\M C),
\end{split}\label{q11}
\end{equation}
are univariate rational functions
in $y$ (Eq.~\eqref{q11} can be derived by expanding $\det(\M C-\alpha \M I)=0$). 
For the sake of simplicity, let us replace $\alpha_{min}$ with $\alpha$.
The necessary condition for minimizing the smallest eigenvalue of $\M C$ is that derivative $\frac{\mathrm{d} \alpha}{\mathrm{d} y}$ should be zero. 
In~\eqref{q10}, $f_1, f_2, f_3$ and $\alpha$ are functions of $y$. Thus, using the assumption $\frac{\mathrm{d} \alpha}{\mathrm{d} y} = 0$, the derivative of~\eqref{q10} gives us
\begin{equation}
\frac{\mathrm{d} f_1}{\mathrm{d} y}\alpha^2-\frac{\mathrm{d} f_2}{\mathrm{d} y}\alpha+\frac{\mathrm{d} f_3}{\mathrm{d} y}=0. \label{q12}
\end{equation}

Rational function $f_1$~\eqref{q11} can be rewritten as $f_1=\frac{g_1}{\delta^2}$, where $g_1$ is a quartic polynomial in $y$ and $\delta=1+y^2$. On the other hand, due to the inner associations in the elements of matrix $\M C$, $f_2$ can be rewritten as $f_2=\frac{g_2}{\delta^3}$, and $f_3$ can be written as $f_3=\frac{g_3}{\delta^4}$, where $g_2, g_3$ are polynomials of degree 6 and 8, respectively. Hence, the derivatives of $f_1,f_2,f_3$ can be rewritten as
\begin{equation}
\frac{\mathrm{d} f_1}{\mathrm{d} y}=\frac{h_1}{\delta^3},\ \frac{\mathrm{d} f_2}{\mathrm{d} y}=\frac{h_2}{\delta^4},\ \frac{\mathrm{d} f_3}{\mathrm{d} y}=\frac{h_3}{\delta^5}, \label{q13}
\end{equation}
where $h_1,h_2,h_3$ are polynomials in $y$ of degree $\{4,6,8\}$, respectively. Substituting the formulas into~\eqref{q10} and~\eqref{q12}, multiplying ~\eqref{q10} by $\delta^4$ and~\eqref{q10} by $\delta^5$, and  substituting  $\beta=\delta \alpha$, we obtain the following two polynomial equations in two unknowns \{$\beta,y$\}.
\begin{equation}
\begin{split}
\delta\beta^3-\beta^2g_1 + \beta g_2 -g_3=0,\\
\beta^2h_1 - \beta h_2 +h_3=0.
\end{split} \label{q15}
\end{equation}
Our objective is to solve these equations efficiently.

\section{Globally Optimal Solver}\label{poly}

A straightforward way of solving the system of two polynomial equations in two unknowns~\eqref{q15} is to use the \gb method~\cite{cox2006using} and generate a specific solver using an automatic generator, \eg,~\cite{kukelova2008automatic,larsson2017efficient,larsson2018beyond}. Using~\cite{larsson2017efficient}, we obtained a \gb solver with a template of size $17\times 45$ for the Gauss-Jordan elimination and 28 solutions. 
However, our experiments on real-world and synthetic data showed that this solver becomes unstable when we are given more than 30 point correspondences, which is almost always the case in real applications. The reason is that the matrix for the Gauss-Jordan elimination is often ill-conditioned. 
Therefore, to find a stable solution, we use the hidden variable technique~\cite{cox2006using} instead.


By treating $y$ as a hidden variable, \ie, by considering $y$ as a coefficient and hiding it into the coefficient matrix, we can rewrite our system~\eqref{q15} as
\begin{equation}
\begin{bmatrix}
-g_3 &  g_2 & - g_1 & \delta  \\
h_3 & - h_2 &  h_1 & 0
\end{bmatrix}\begin{bmatrix}
1  \\
\beta \\
\beta^2 \\
\beta^3
\end{bmatrix}=0,\label{q16}
\end{equation}
where $g_1,\dots,g_3,h_1,\dots,h_3$ and $\delta$ are polynomials in $y$.
With this formulation we have two equations with 4 monomials. Therefore, to easily solve this system by ``linearizing'' it, we need to add additional equations to~\eqref{q16} to obtain as many equations as monomials. In our case, we only need to multiply the first equation by $\beta$, and the second equation by $\{\beta,\beta^2\}$. In this way, we obtain 5 equations with 5 monomials, which can be written as 
\begin{equation}
{\M B(y)} \boldsymbol\omega=\M 0,\label{q17}
\end{equation}
where $\boldsymbol\omega=[1\ \beta\ \beta^2\ \beta^3\ \beta^4]^\top$ and $\M B(y)$ is the $5\times5$  matrix
\begin{equation}
\M B(y)= \begin{bmatrix}
-g_3 & g_2 & -g_1 & \delta & 0 \\
h_3 & - h_2 & h_1 & 0 & 0 \\
0& -g_3 & g_2 & - g_1 & \delta \\
0& h_3 & - h_2 &  h_1 & 0  \\
0&  0& h_3 & - h_2 & h_1  
\end{bmatrix},\label{q18}
\end{equation}
whose elements are polynomials in $y$. Next, we will present two solutions to~\eqref{q17}.

\vspace{1mm}
\noindent\textbf{Sturm Sequence Solution.} 
Since the matrix $\M B(y)$ has a right null vector, the determinant of $\M B(y)$ must vanish. The sparse structure of $\M B(y)$ allows us to use the Laplace expansion to obtain $\det(\M B(y))$, which is a univariate polynomial in $y$ of degree 28. 
Then, the Sturm sequence method is used to find the real roots of the obtained univariate polynomial, which is fast and efficient. For more details about the Sturm root bracketing, we refer the reader to~\cite{gellert2012vnr}.

\vspace{1mm}
\noindent\textbf{Polynomial Eigenvalue Solution.}
Note, that~\eqref{q17} is essentially a polynomial eigenvalue problem (PEP) of degree 8, which can be written as
\begin{equation}
(y^8{\M B}_8+y^7{\M B}_7+...+{\M B}_0)\ \boldsymbol\omega =0,\label{q19}
\end{equation}
where ${\M B}_8,{\M B}_7,...,{\M B}_0$ are $5\times5$ coefficient matrices. 
PEP~\eqref{q19} can be transformed to an eigenvalue problem
$
\M D  \boldsymbol\psi = z \boldsymbol\psi,\label{q20}
$
with $z = 1/y$, $\boldsymbol\psi = [\boldsymbol\omega, z \boldsymbol\omega,\dots, z^7\boldsymbol\omega]^\top$,  and $40 \times 40$ matrix $\M D$ of the form
\begin{equation}
{\M D}=\begin{bmatrix}
\vspace{1mm}
\M 0 & {\M I}&  ...& \M 0\\
\vspace{1mm}
\M 0 & \M 0&  ...& \M 0\\
\vspace{1mm}
... & ...&  ...& \M I\\
\vspace{1mm}
-{\M B_0^{-1}}{\M B_8} & -{\M B_0^{-1}}{\M B_7} &  ... & -{\M B_0^{-1}}{\M B_1} 
\end{bmatrix}.\label{23}
\end{equation}
Note, that this eigenvalue formulation is a relaxation of the original problem~\eqref{q16} that does not consider the monomial dependences in $\boldsymbol\psi$, and therefore, it introduces some spurious solutions.
Six of these spurious solutions can be, however, easily removed.
These solutions correspond to zero eigenvalues of matrix $\M D$ that are generated by zero columns in matrices ${\M B_2}$ and ${\M B_1}$.
After removing these columns and corresponding rows, the size of  ${\M D}$ reduces to $34\times 34$. 
The real eigenvalues of $\M D$, which can be computed from real Schur decomposition, are the solutions to $z = 1/y$.
Note, that we don't need to compute the eigenvectors of the $40\times 40 $ matrix $\M D$ since $\alpha_{min}$~\eqref{q9} and the translation vector can be extracted from the eigenvalues and eigenvectors of the $3\times3$ matrix $\M C$.
This polynomial eigevalue solution is slower than the solution based on Sturm sequences, however, as we will show in experiments, it is more stable.

\section{Linearized Solver}

In visual odometry and SLAM applications, the relative rotation between consecutive frames is often small or negligible. 
Therefore, the first-order Taylor expansion usually leads to a reasonably good approximation for rotation.
In this case, the rotation matrix can be simplified as
\begin{equation}
\M R_y=\begin{bmatrix}
1 & 0 & \theta\\
0 & 1 & 0\\
- \theta & 0 & 1
\end{bmatrix},\label{eq:linrot}
\end{equation}
and the elements of matrix $\M C$ \eqref{q8a} are quadratic polynomials in $\theta$. Similar to~\eqref{q15}, we may obtain two polynomials with respect to $\{\theta,\alpha\}$ as follows:
\begin{equation}
\begin{split}
\alpha^3-f_1\alpha^2+f_2\alpha-f_3=0,\\
g_1\alpha^2-g_2\alpha+g_3=0,
\end{split} \label{eq:lintwo}
\end{equation}
where $f_1,f_2,f_3$ are polynomials in $\theta$ of degree $\{2,4,6\}$, and $g_1 = \frac{\mathrm{d} f_1}{\mathrm{d} \theta}, g_2 = \frac{\mathrm{d} f_2}{\mathrm{d} \theta}, g_3 = \frac{\mathrm{d} f_3}{\mathrm{d} \theta}$ are polynomials in $\theta$ of degree $\{1,3,5\}$, respectively. 
This formulation is simpler than~\eqref{q15}, since the polynomials have lower degrees. This system has up to $15$ real solutions and the \gb solver generated using~\cite{larsson2017efficient} has a template matrix of size $15\times30$. 
This solver is, however, unstable. Therefore, we use the Sturm sequence and the polynomial eigenvalue technique~\cite{kukelova2012polynomial} as for the non-linearized solver to generate a efficient and stable solvers for \eqref{eq:linrot}. The main steps are the same as in Section~\ref{poly}. The only difference is that we only need to find the roots of a univariate polynomial in degree 15 or eigenvalues of a $21\times 21$ matrix.

\section{Experiments}\label{exp}

We studied the performance of the proposed PEP-based optimal (OPT) and linearized (LIN) solvers and their variants based on Sturm sequences (OPT-S) and (LIN-S) on synthetic and real-world images. 
In the comparison, we included the closely related work of Ding et al.\ (3PC)~\cite{ding_2020icra} and the globally optimal SDP solution (SDP)~\cite{zhao2020pami}.
In addition, we compared the proposed solver with the 5PC algorithm~\cite{stewenius2006recent} and the normalized 8PC solver~\cite{hartley1995defence} with and without a final Levenberg-Marquardt~\cite{more1978levenberg} numerical optimization (8PC+LM). We did not include 3PC+LM and 5PC+LM since their performance was similar to 8PC+LM.

\subsection{Synthetic Evaluation}

We chose the following setup to generate synthetic data. 
First, 200 random 3D points and 200 image pairs with random poses were generated. 
The focal length $f_g$ of the camera was set to 1000 pixels. 
The parameters which were changed to test the performance are the noise level ($\sigma$) in the image point locations, the baseline ($\M t$) between two cameras, the number of points ($N$) used for the solvers, the field of view (FOV) of the camera, and the noise level $\tau$ in the gravity direction. 
The default setting: $\sigma$ = 1 pixel, $\M t$ = 5\% of the average scene depth, $N$ = 20, FOV = 90$^\circ$, and $\tau$ = 0$^\circ$. It is reasonable to assume that we have an almost perfect measurement of the gravity direction, since we obtain the gravity vector by applying a high-pass filter to isolate the force of gravity from the raw accelerometer data. 
The performance of the solvers was tested by modifying the value of a single parameter from the aforementioned ones. 
The rotation error was defined as the angle difference between the estimated rotation and the ground truth rotation as ${\rm arccos}\left(\left({\rm tr}(\M R_g\M R^{\top}_e)-1\right)/2\right)$, where $\M R_g$ and $\M R_e$ are the ground truth and estimated rotations, respectively. The translation error was measured as the angle between the estimated and ground truth translation vectors, since the estimated translation is recovered only up to scale. 

\begin{figure*}
	\centering
	\subfloat[]{\label{figure:a1} \includegraphics[trim={2mm 0mm 1mm 0mm},clip,width=1.64in]{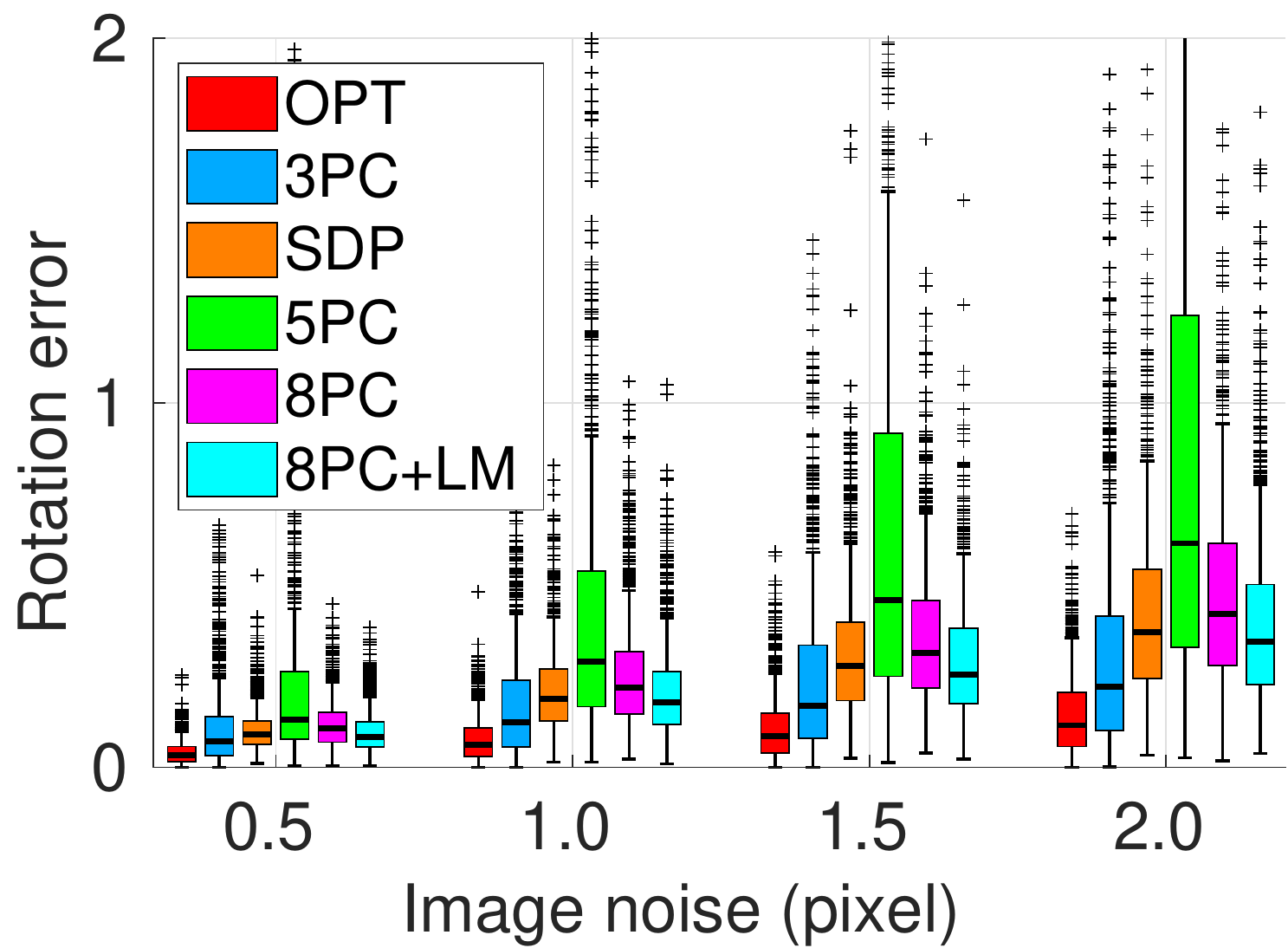}}\,
	\subfloat[]{\label{figure:a2} \includegraphics[trim={2mm 0mm 1mm 0mm},clip,width=1.64in]{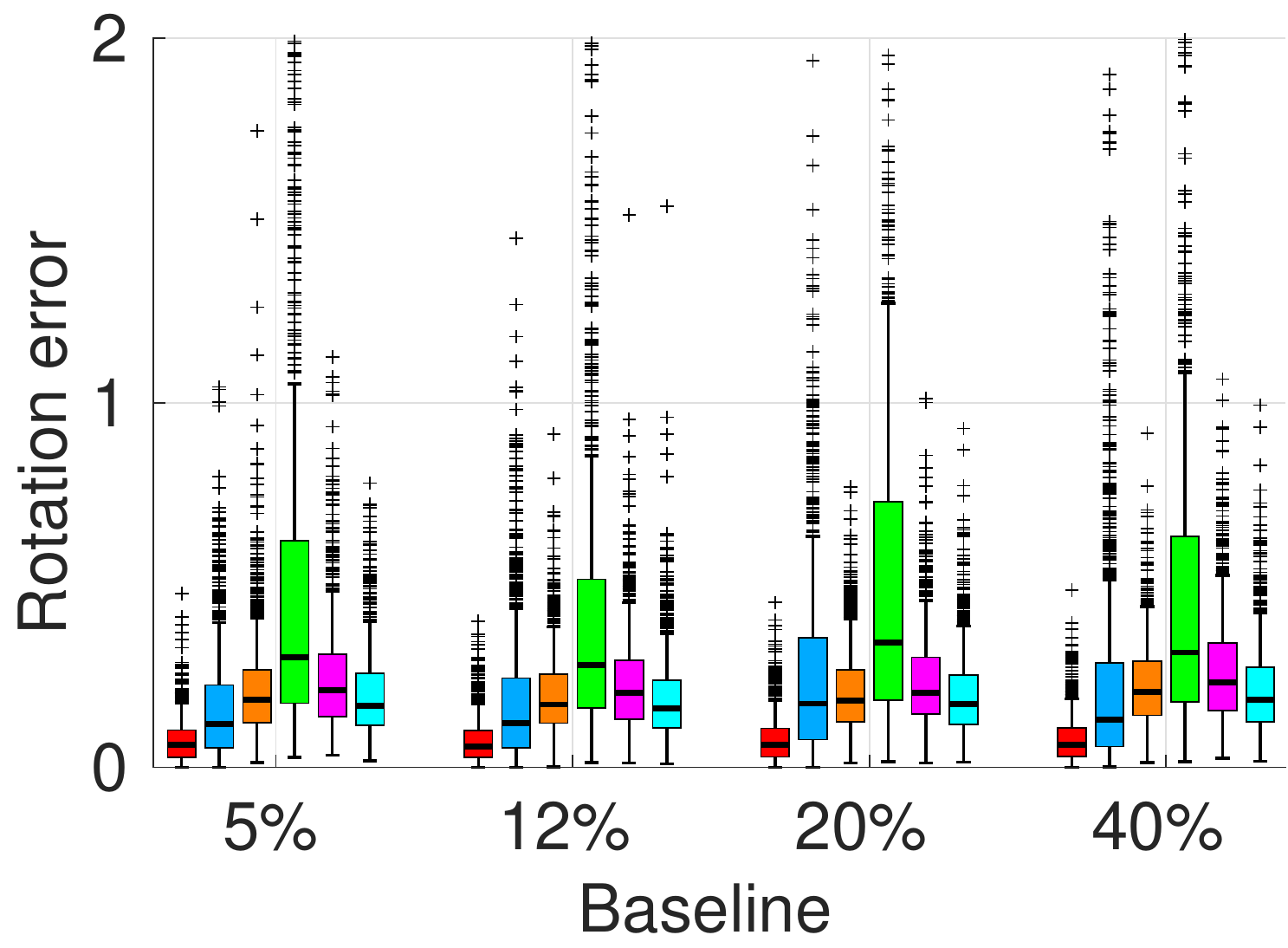}}\,
	\subfloat[]{\label{figure:a3} \includegraphics[trim={2mm 0mm 1mm 0mm},clip,width=1.64in]{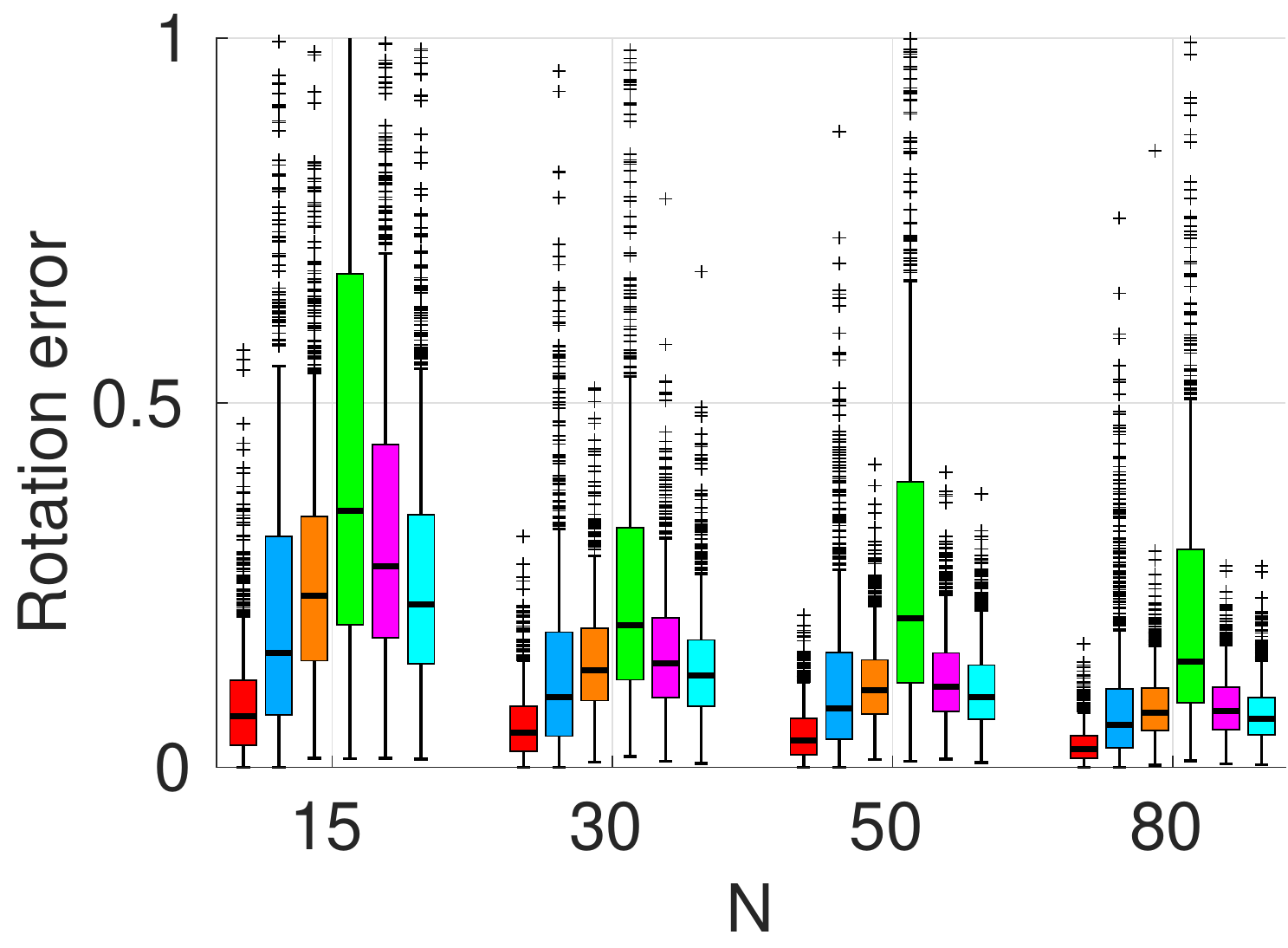}}\,
	\subfloat[]{\label{figure:a4} \includegraphics[trim={2mm 0mm 1mm 0mm},clip,width=1.64in]{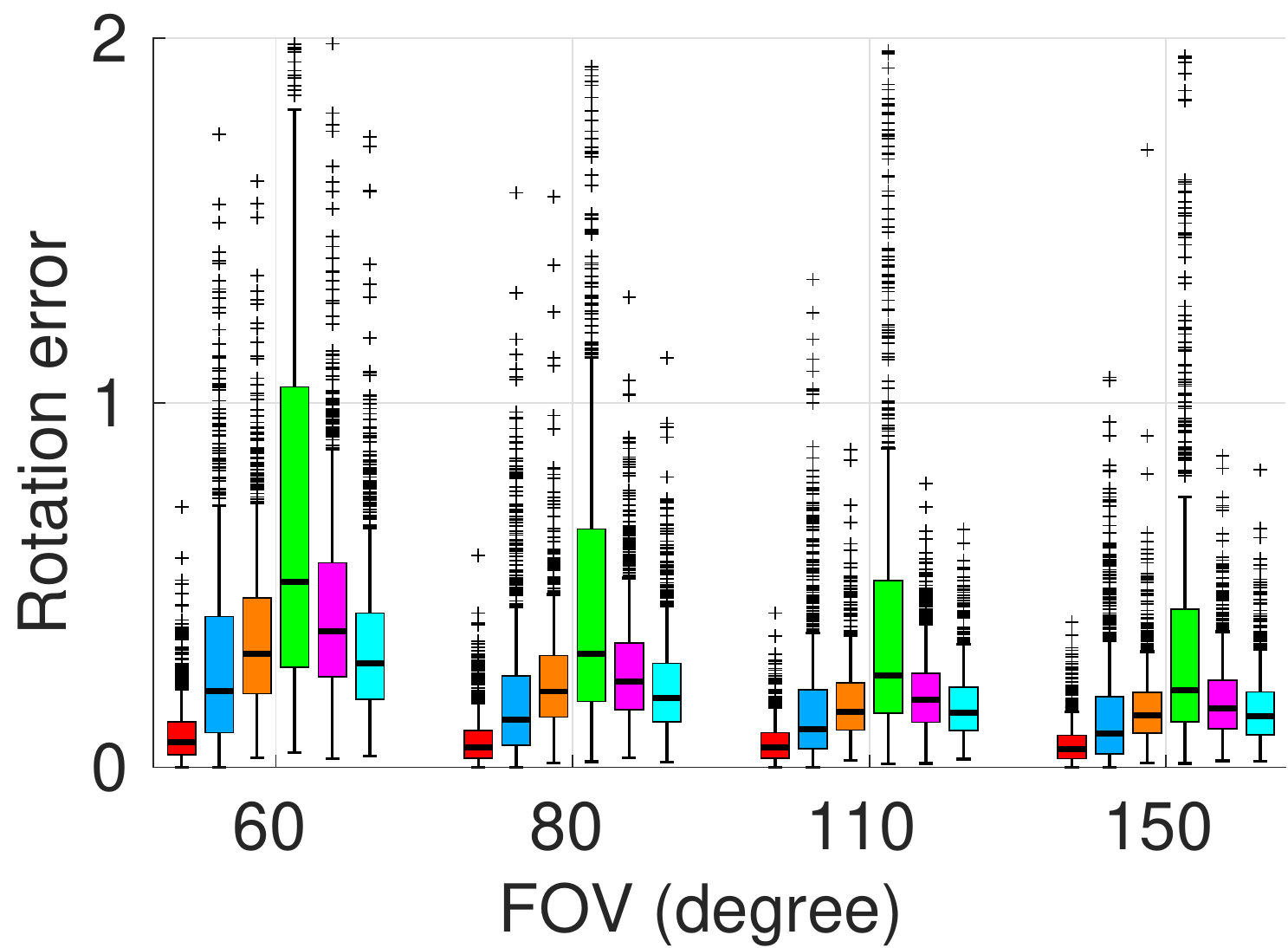}}\\
	\subfloat[]{\label{figure:b1} \includegraphics[trim={2mm 0mm 1mm 0mm},clip,width=1.64in]{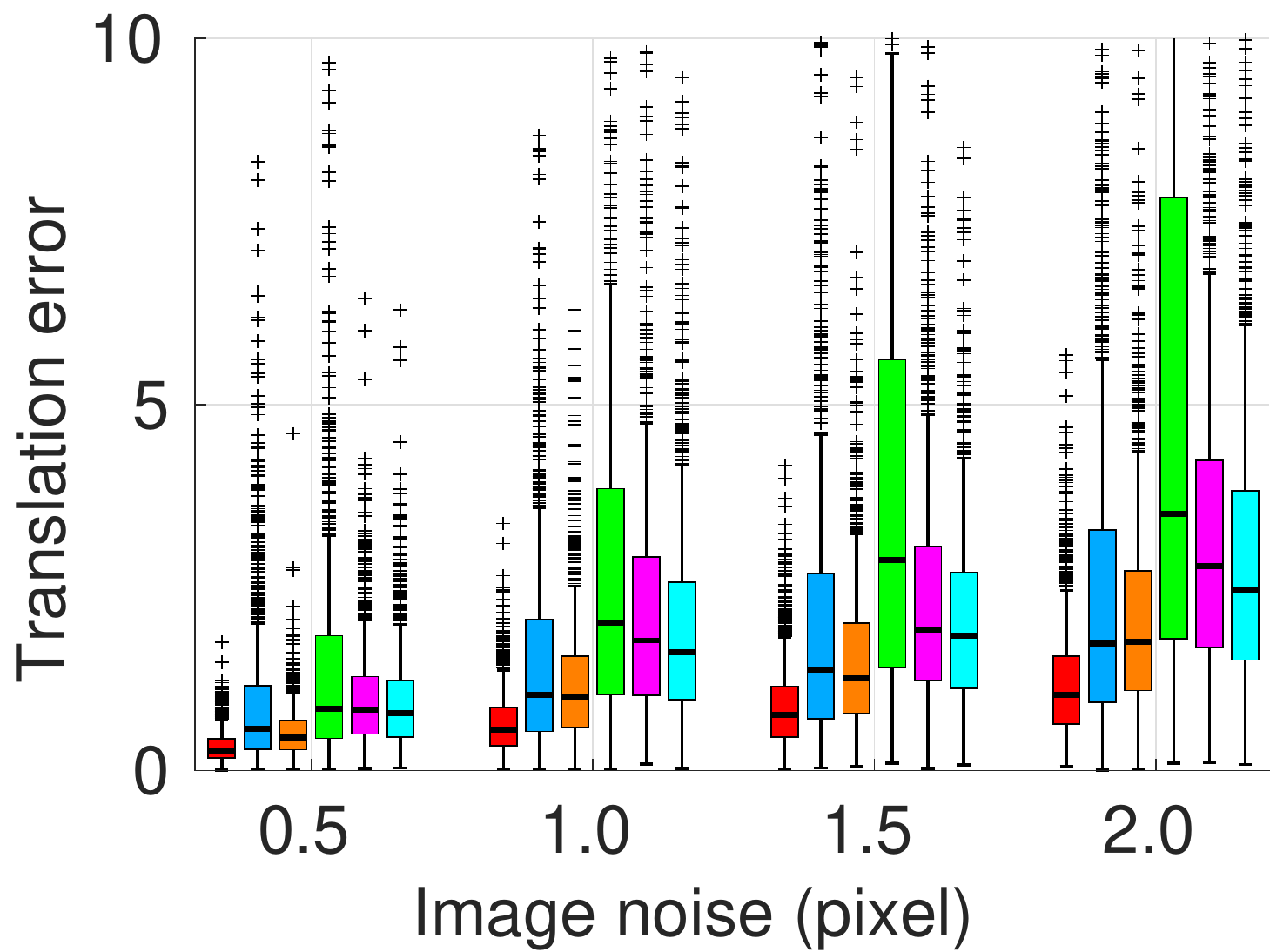}}\,
	\subfloat[]{\label{figure:b2} \includegraphics[trim={2mm 0mm 1mm 0mm},clip,width=1.64in]{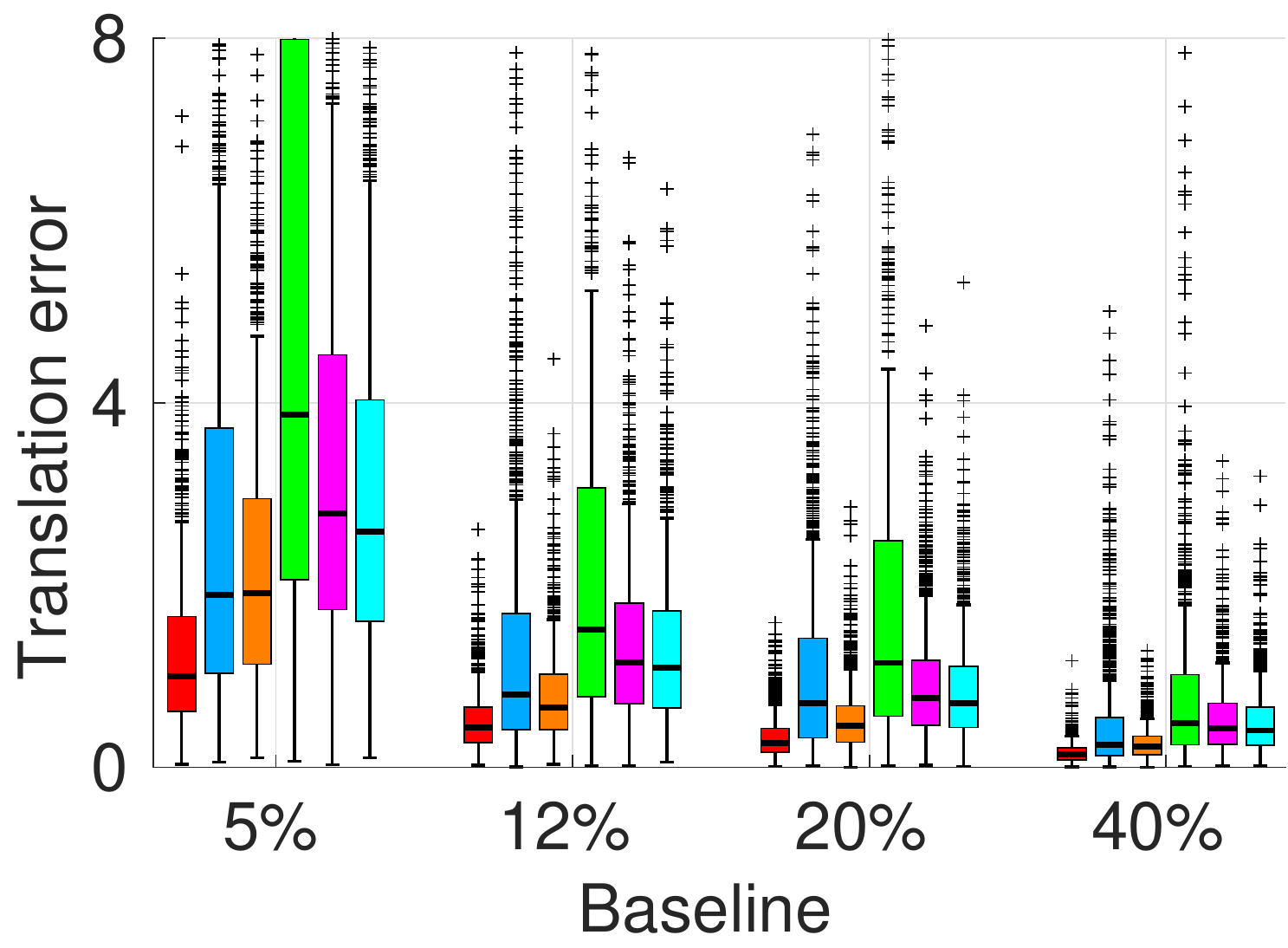}}\,
	\subfloat[]{\label{figure:b3} \includegraphics[trim={2mm 0mm 1mm 0mm},clip,width=1.64in]{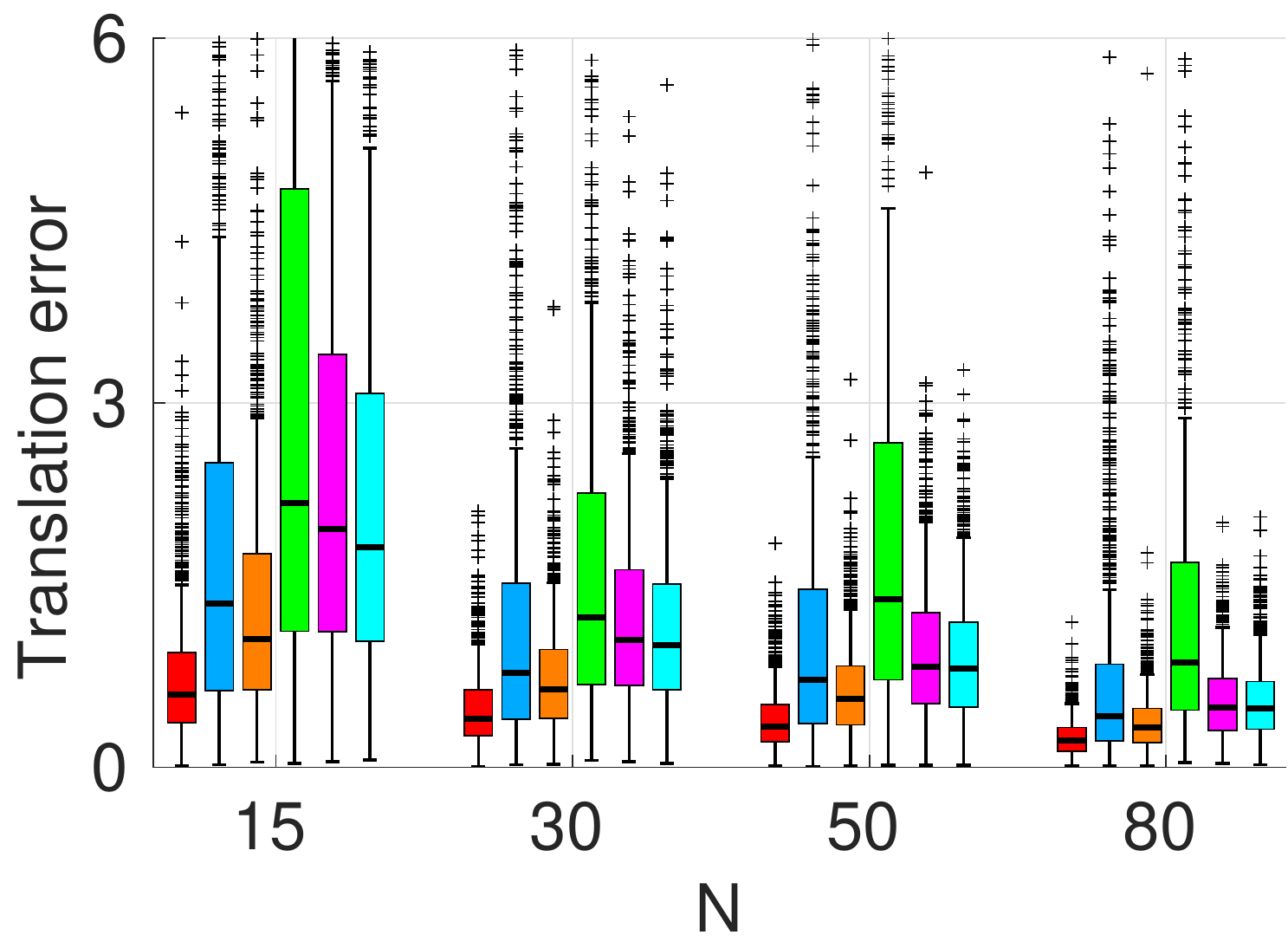}}\,
	\subfloat[]{\label{figure:b4} \includegraphics[trim={2mm 0mm 1mm 0mm},clip,width=1.64in]{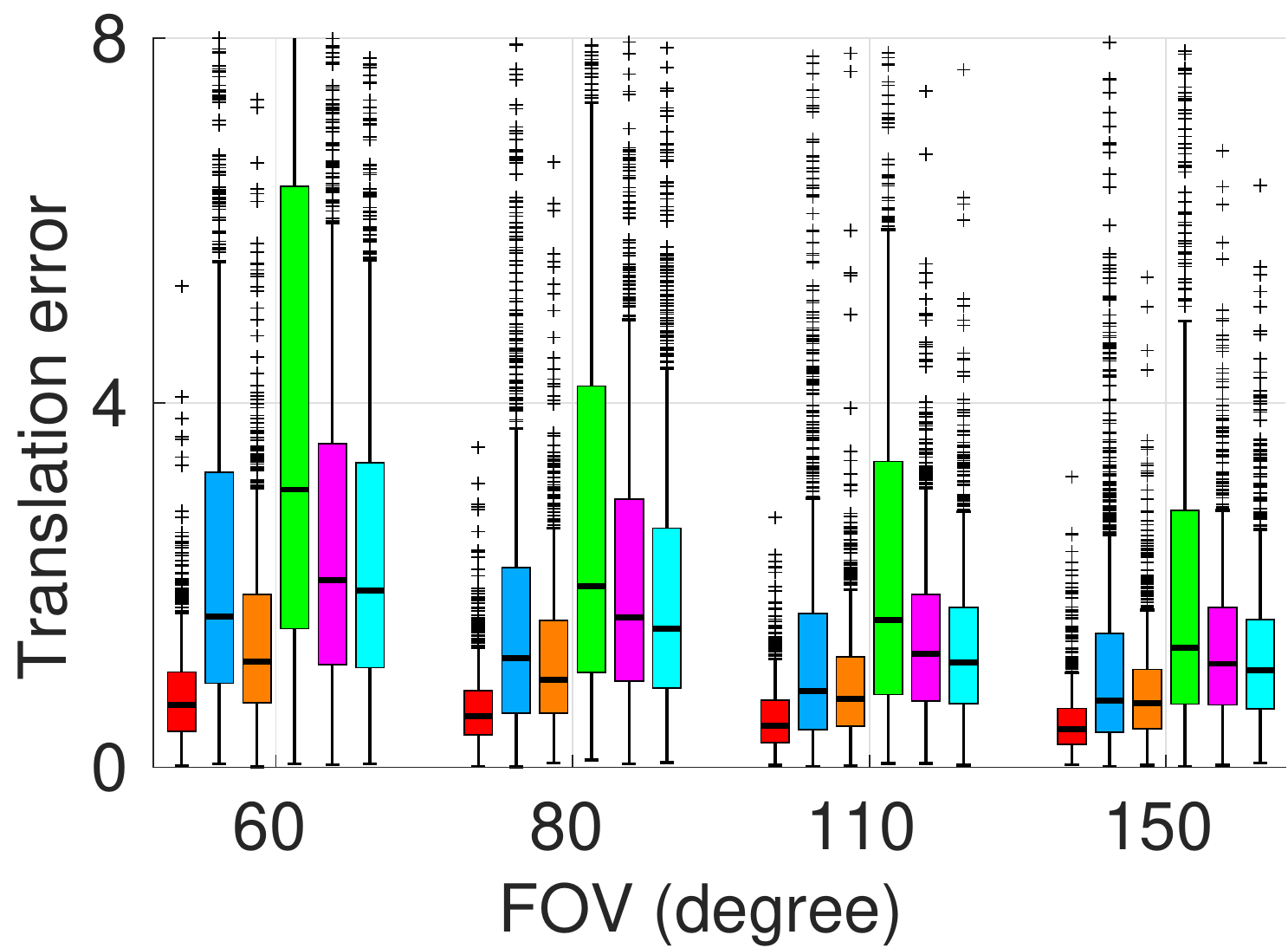}}
	\caption{
	\textbf{Top}: rotation error in degrees. 
	\textbf{Bottom}: translation error in degrees. 
	The columns show the error of the solvers against increasing (a) image noise, (b) baseline, (c) correspondence number, and (d) field-of-view.  
	We use the following default values for the parameters not tested in a figure: std.\ of the image noise is $\sigma$ = 1 px; length of the baseline = 10\% scene depth; number of correspondences N = 20; field of view FOV = 90$^\circ$, the std.\ of the gravity directional noise $\tau$ = 0$^\circ$.}
	\label{syn:pixel}
\end{figure*}

\begin{figure}
	\centering
	\subfloat[]{\label{figure:roll_pitchr} \includegraphics[trim={2mm 0mm 1mm 0mm},clip,width=1.57in]{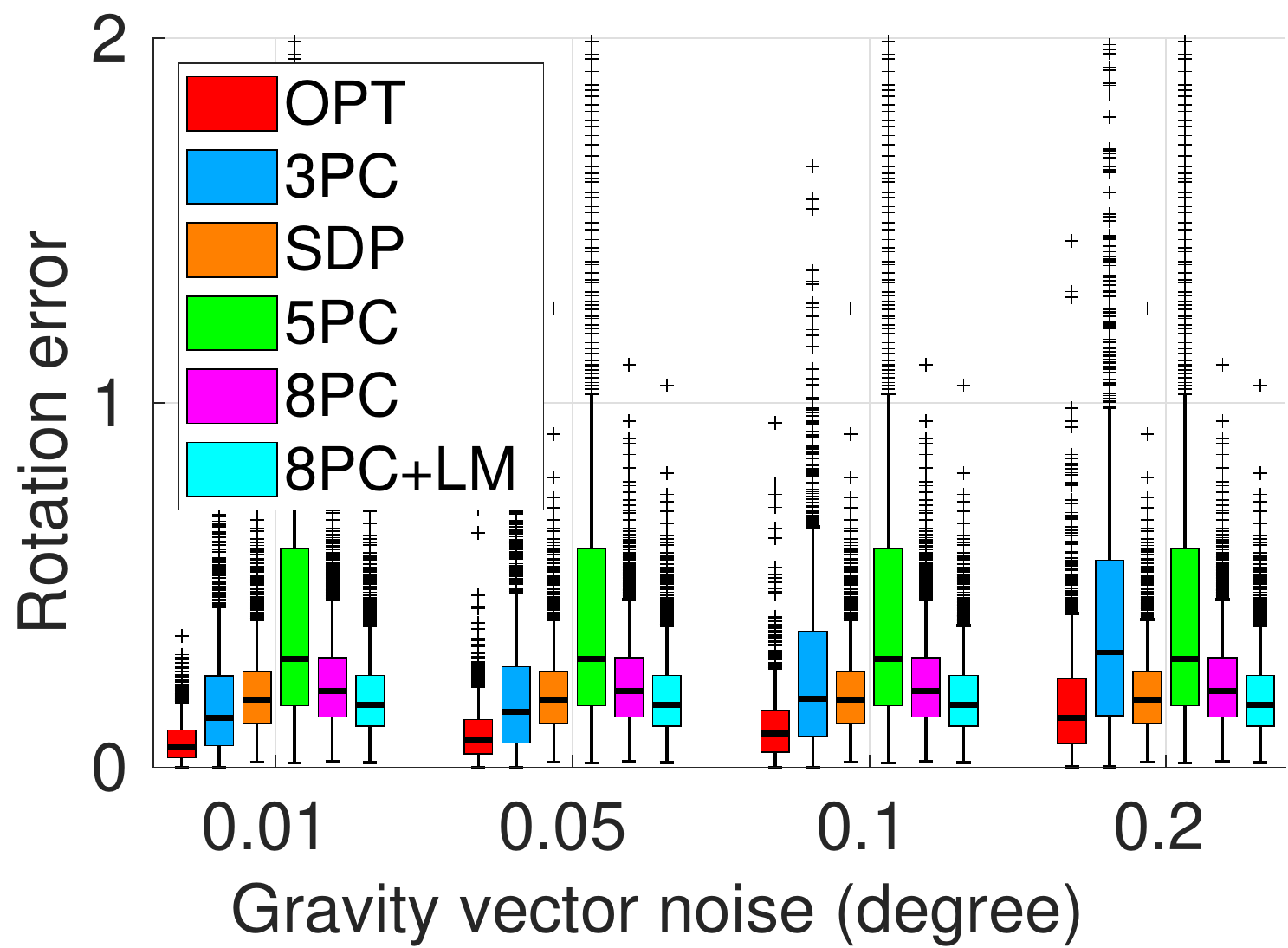}}\,
	\subfloat[]{\label{figure:roll_pitcht} \includegraphics[trim={2mm 0mm 1mm 0mm},clip,width=1.57in]{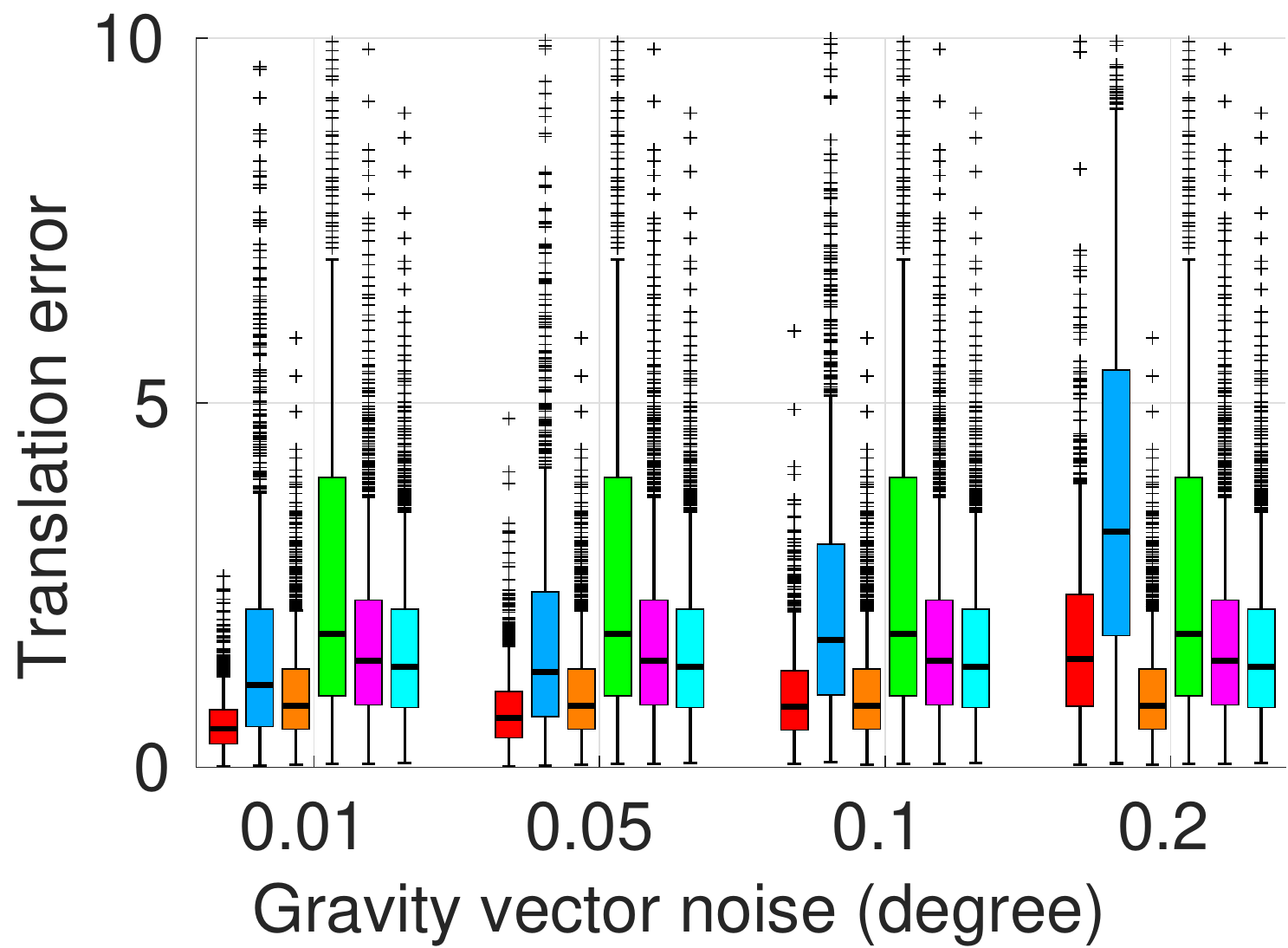}}
	\caption{Rotation and translation errors in degrees. Default settings with increased roll and pitch noise.}
	\label{syn:gravity}
\end{figure}

\paragraph{Solver accuracy.}
Fig.~\ref{syn:pixel} and Fig.~\ref{syn:gravity} report the rotation (top row) and translation (bottom) errors under different configurations. From Fig.~\ref{syn:pixel}, we can see that the proposed optimal solver (OPT) outperforms all the existing methods in the case when the gravity direction is perfect. However, in real applications, the gravity vector isolated from the raw accelerometer data may not be perfect, and we also studied the performance under increased gravity direction noise. Since accelerometers used in cars and modern smartphones have noise levels around 0.06$^\circ$ (or an expensive “good” accelerometers have less than 0.02$^\circ$), we added noise to roll and pitch angles for both views with a maximum value of 0.2$^\circ$. Fig.~\ref{syn:gravity} shows that our OPT solver is still comparable to the SOTA even when the noise level is up to 0.2$^\circ$. Due to the lack of space and for batter readability of graphs we do not include results of OPT-S, LIN and LIN-S solvers here. The results of these solvers, including the results for increasing relative rotation, forward  motion, and stability experiments are in the supplementary material.

\begin{figure}[t]
    \centering
	\includegraphics[width=0.75\columnwidth]{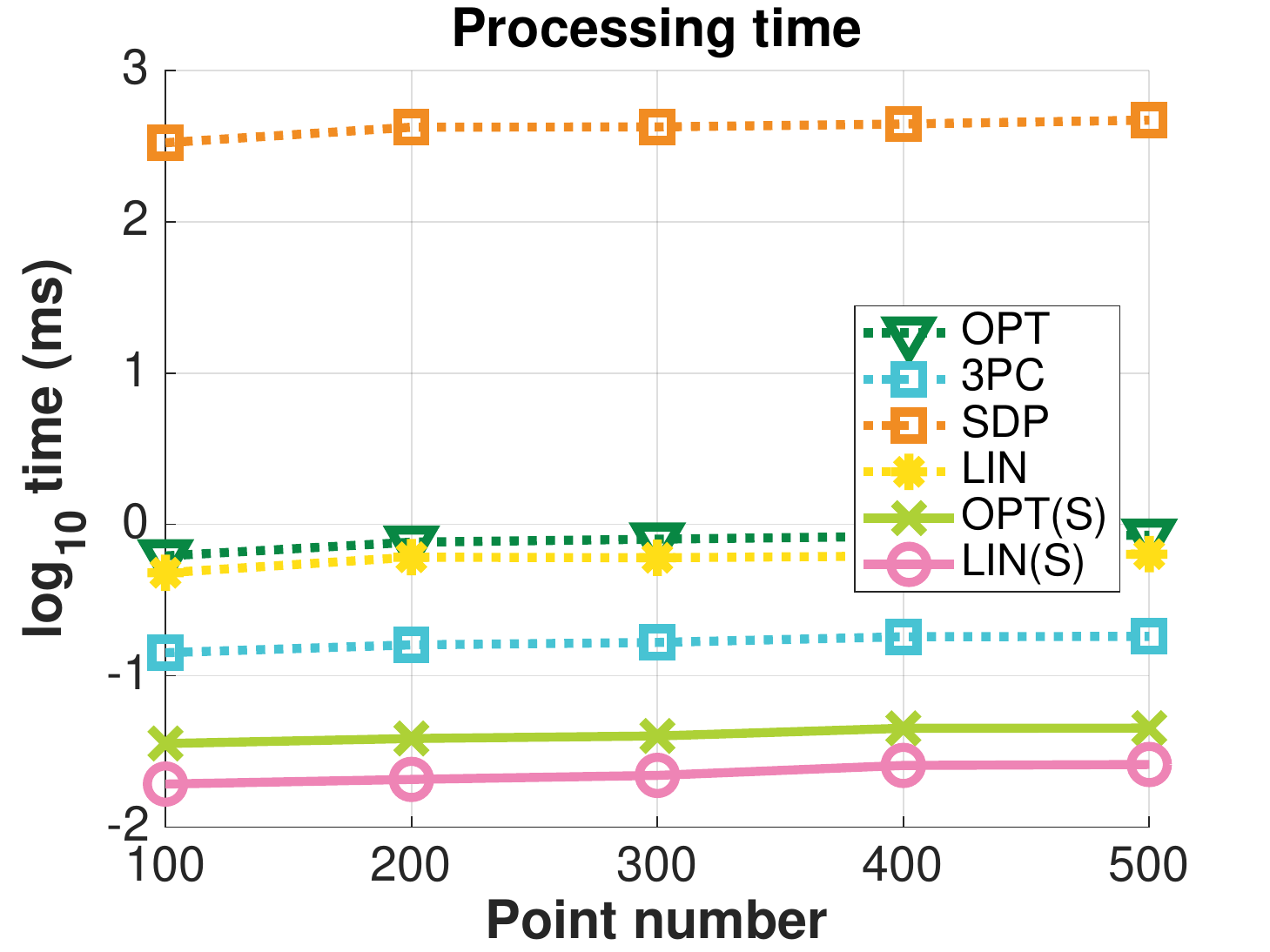}
    \caption{The $\log_{10}$ processing time in milliseconds plotted as the function of the point number used for the estimation.}
    \label{fig:solver_times}
\end{figure}

\paragraph{Complexity analysis and running times.}
The processing times of the solvers are shown in Fig.~\ref{fig:solver_times}. The proposed optimal (OPT) and linearized (LIN) ones
are almost three orders of magnitude faster than the SDP optimal solver of~\cite{zhao2020pami}.
Note that, for~\cite{zhao2020pami}, we used an alternative implementation based on the cvx toolbox~\cite{grant2014cvx}. Our implementation has an average of $200ms$ compared to~\cite{zhao2020pami} which has an average of $6ms$ based on the sdpa toolbox according to the paper.
The solvers based on Sturm sequences are almost and order of magnitude faster than the minimal 3PC solver~\cite{ding_2020icra}.
%
The following table contains the main operations performed by the proposed solvers, together with their running times. The second column shows the size of the matrix for the Gauss-Jordan elimination, the third column shows the size of the matrix for the eigenvalue decomposition, and the fourth one shows the degree of the univariate polynomial solved by Sturm sequences.
\begin{table}[h]
	\newcommand{\tabincell}[2]{\begin{tabular}{@{}#1@{}}#2\end{tabular}}
	\begin{center}
		\begin{tabular}{lcccc}
			\toprule
			Solver & G-J & Eigen & Sturm & Time ($\mu$s)  \\
			\midrule
			OPT & $5\times34$ & $34\times34$ &  - & 115  \\ 
			OPT-S & - & - & 28 & 24  \\ 
			LIN & $5\times21$ & $21\times21$ & - & 54\\
			LIN-S & - & - & 15 & 13  \\
			\bottomrule
		\end{tabular}
	\end{center}
	\vspace{-1.2em}
	\label{tab:complex}
\end{table}

\subsection{Rear-world Experiments}

In order to test the proposed technique on real-world data, we chose the \Malaga~\cite{Claraco2014}\footnote{\url{https://www.mrpt.org/MalagaUrbanDataset}}, \KITTI~\cite{Geiger2012CVPR}\footnote{\url{http://www.cvlibs.net/datasets/kitti}} and \ETH~\cite{schops2017multi}\footnote{\url{https://www.eth3d.net/}} datasets. 
\Malaga was gathered entirely in urban scenarios with car-mounted sensors, including one high-resolution stereo camera and five laser scanners. 
We used the sequences of one high-resolution camera and every $10$th frame from each sequence. 
The proposed solvers were applied to every consecutive image pair. 
The ground truth paths were composed using the GPS coordinates provided in the dataset.
In total, $9,064$ image pairs were used from the dataset.
The \KITTI odometry benchmark consists of 22 stereo sequences. Only 11 sequences (00--10) are provided with ground truth trajectories for training. 
We therefore used these 11 sequences to evaluate the compared solvers. 
In total, $23190$ image pairs were used.
The \ETH dataset covers a variety of indoor and outdoor scenes.
Ground truth geometry has been obtained using a high-precision laser scanner.
A DSLR camera as well as a synchronized multi-camera rig with varying field-of-view was used to capture images. In total, we used $5162$ image pairs.

%
%
For testing non-minimal solvers on real-world data, we chose a locally optimized RANSAC, {\ie},\ Graph-Cut RANSAC\footnote{\url{https://github.com/danini/graph-cut-ransac}}~\cite{barath2017graph} (GC-RANSAC). 
In GC-RANSAC (and other locally optimized RANSACs), two different solvers are used: (a) one for estimating the pose from a minimal sample and (b) one for fitting to a larger-than-minimal sample when doing final pose polishing on all inliers or in the local optimization step. 
For (a), the main objective is to solve the problem using as few correspondences as possible since the processing time depends exponentially on the number of correspondences required for the pose estimation. 
We tested two minimal solvers, the 5PC algorithm of Stewenius et al.\ \cite{stewenius2005minimal} and the 3PC method of Ding et al.\ \cite{ding_2020icra} exploiting the gravity direction similarly to our method.
The purpose of (b) is to estimate the pose parameters as accurately as possible. 
In (b), we tested the proposed solvers, the normalized 8PC algorithm~\cite{hartley1995defence}, the 5PC solver~\cite{stewenius2005minimal}, and the 3PC method of Ding et al.\ \cite{ding_2020icra}.
We excluded \cite{zhao2020pami} from the real-world tests since it is extremely slow, even when implemented in C++, as reported in Fig.~\ref{fig:solver_times}. 

\begin{figure*}[t]
    \centering
    \includegraphics[width=0.8\textwidth]{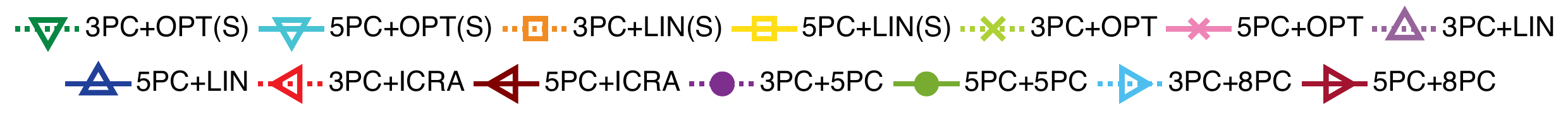}\\
	\subfloat[\Malaga]{\label{figure:a51} 
	    \begin{tabular}[b]{c}
	        \includegraphics[trim={1mm 0mm 10mm 0mm},clip, width=0.295\textwidth]{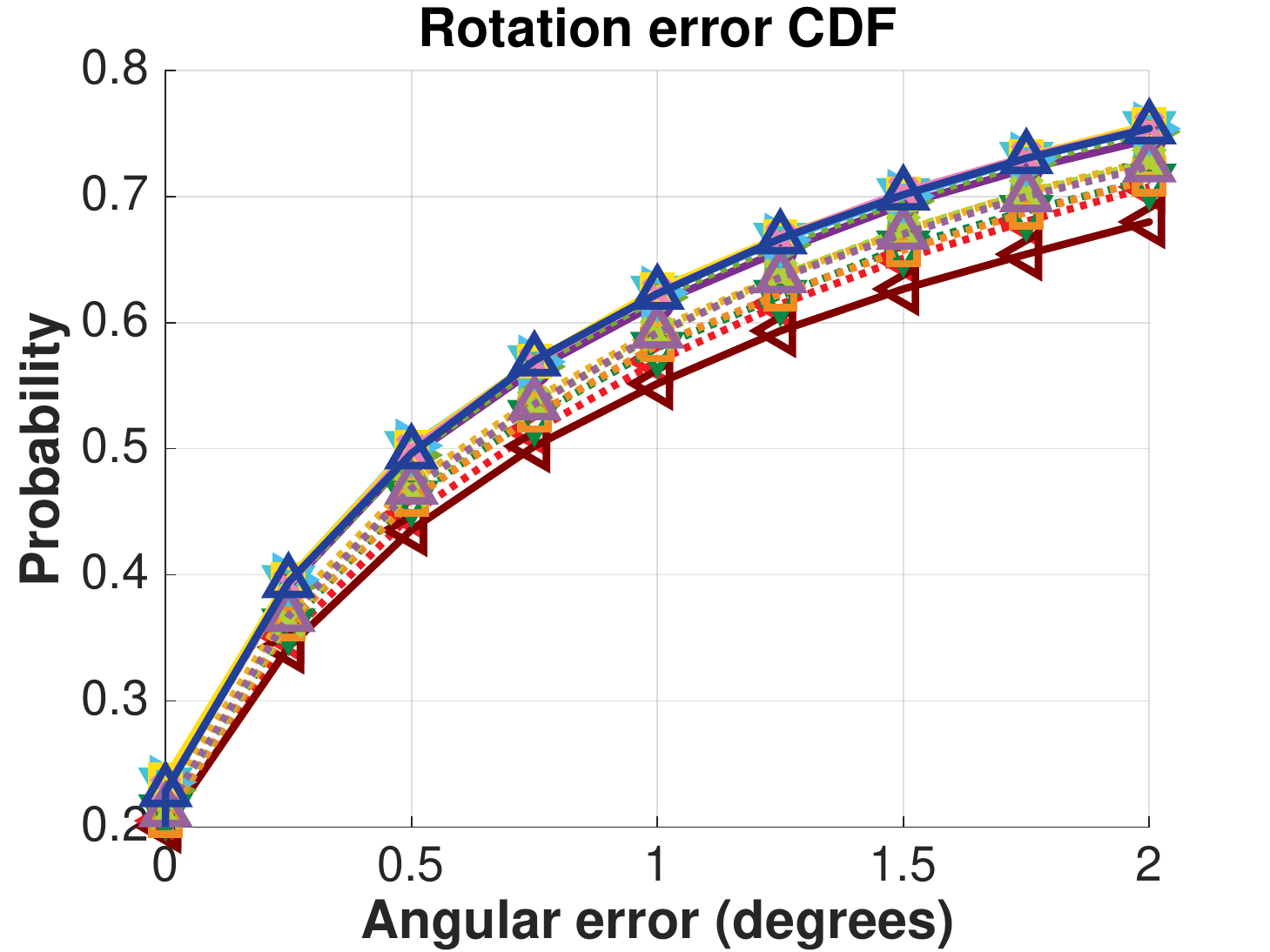}\\ 
	        \includegraphics[trim={1mm 0mm 10mm 0mm},clip, width=0.295\textwidth]{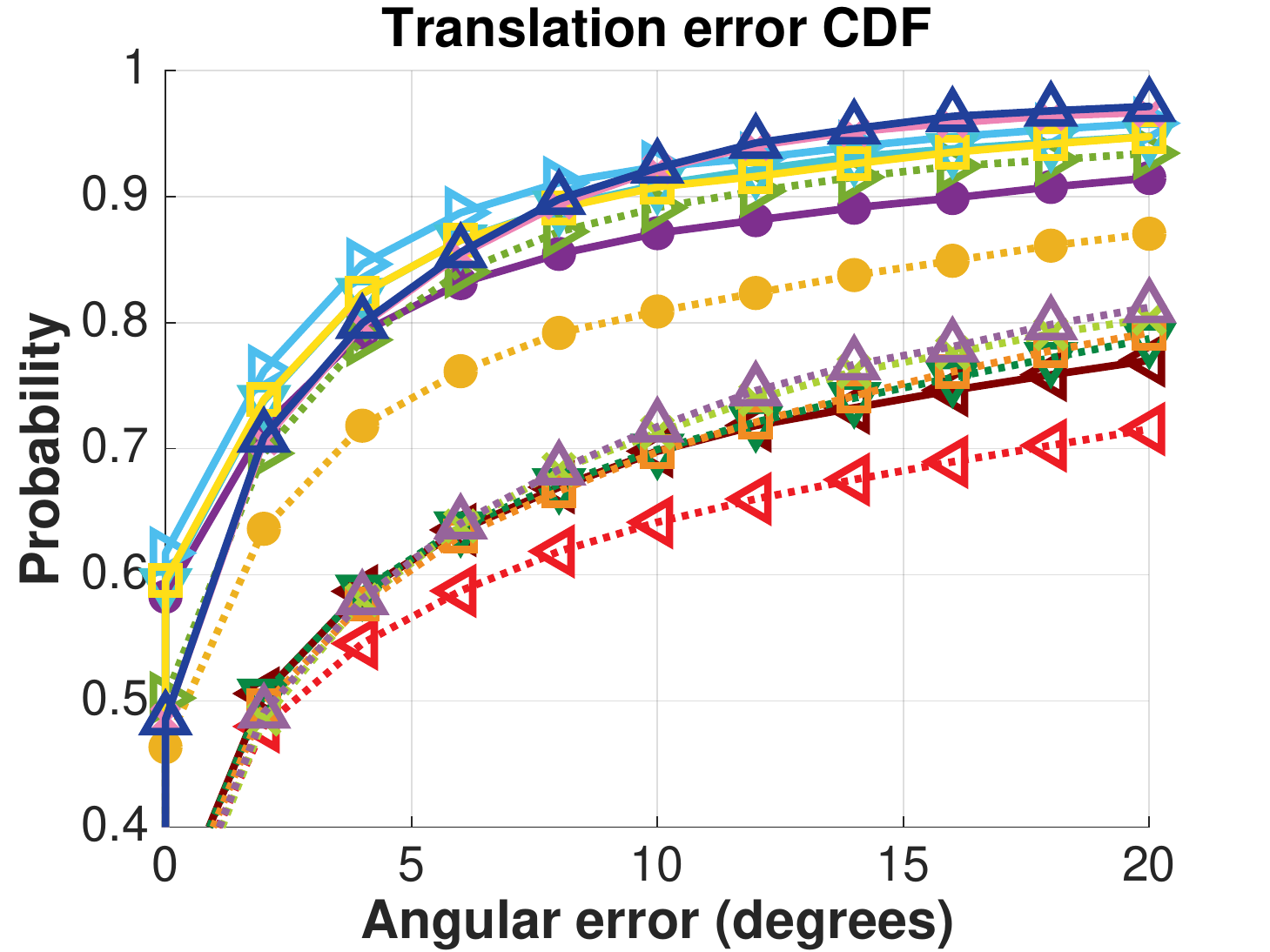}
	    \end{tabular}}
	\subfloat[\KITTI]{\label{figure:a52} 
	    \begin{tabular}[b]{c}
	        \includegraphics[trim={1mm 0mm 10mm 0mm},clip, width=0.295\textwidth]{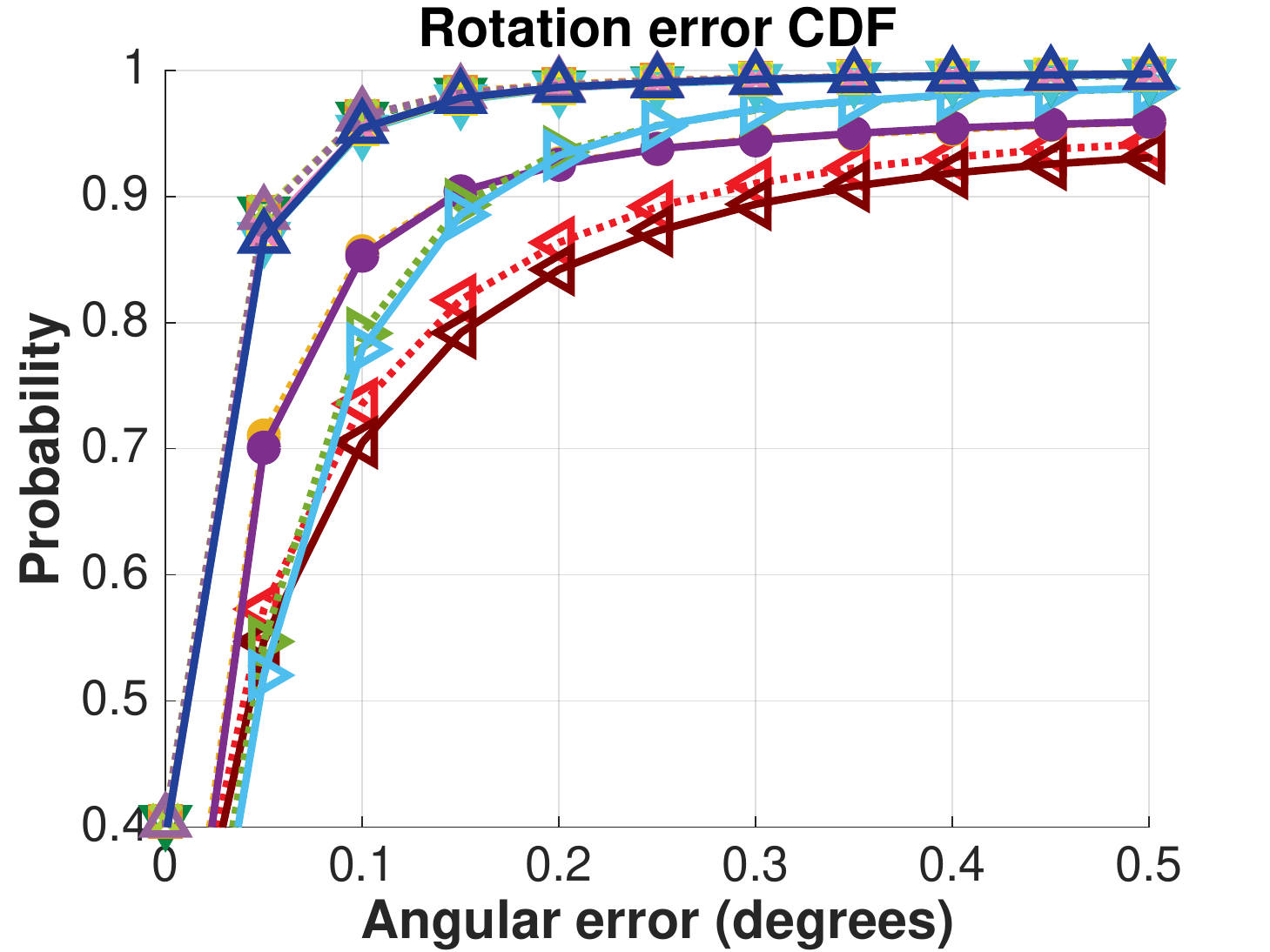}\\ 
	        \includegraphics[trim={1mm 0mm 10mm 0mm},clip, width=0.295\textwidth]{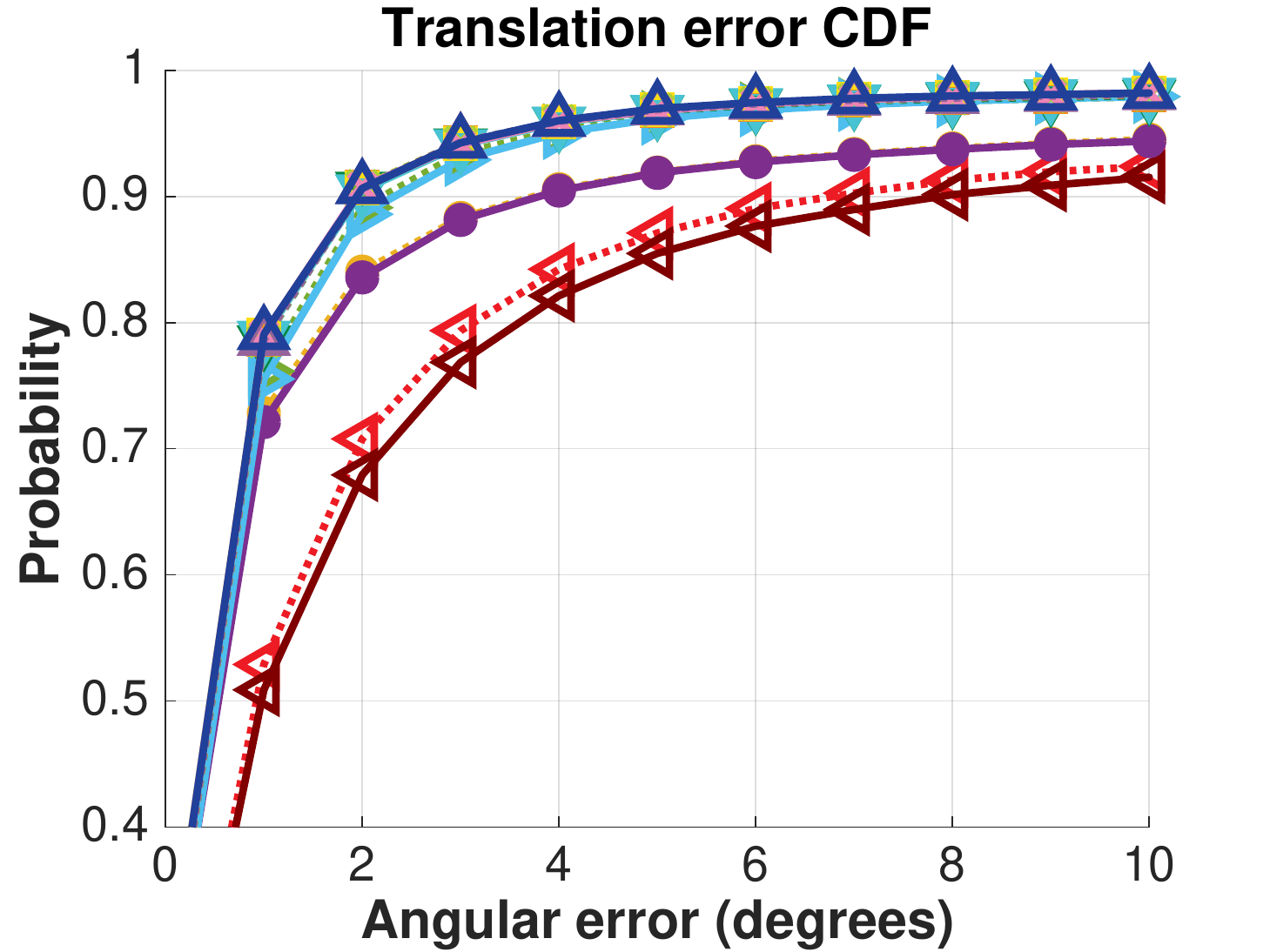}
	    \end{tabular}}
	\subfloat[\ETH]{\label{figure:a53} 
	    \begin{tabular}[b]{c}
	        \includegraphics[trim={1mm 0mm 10mm 0mm},clip, width=0.295\textwidth]{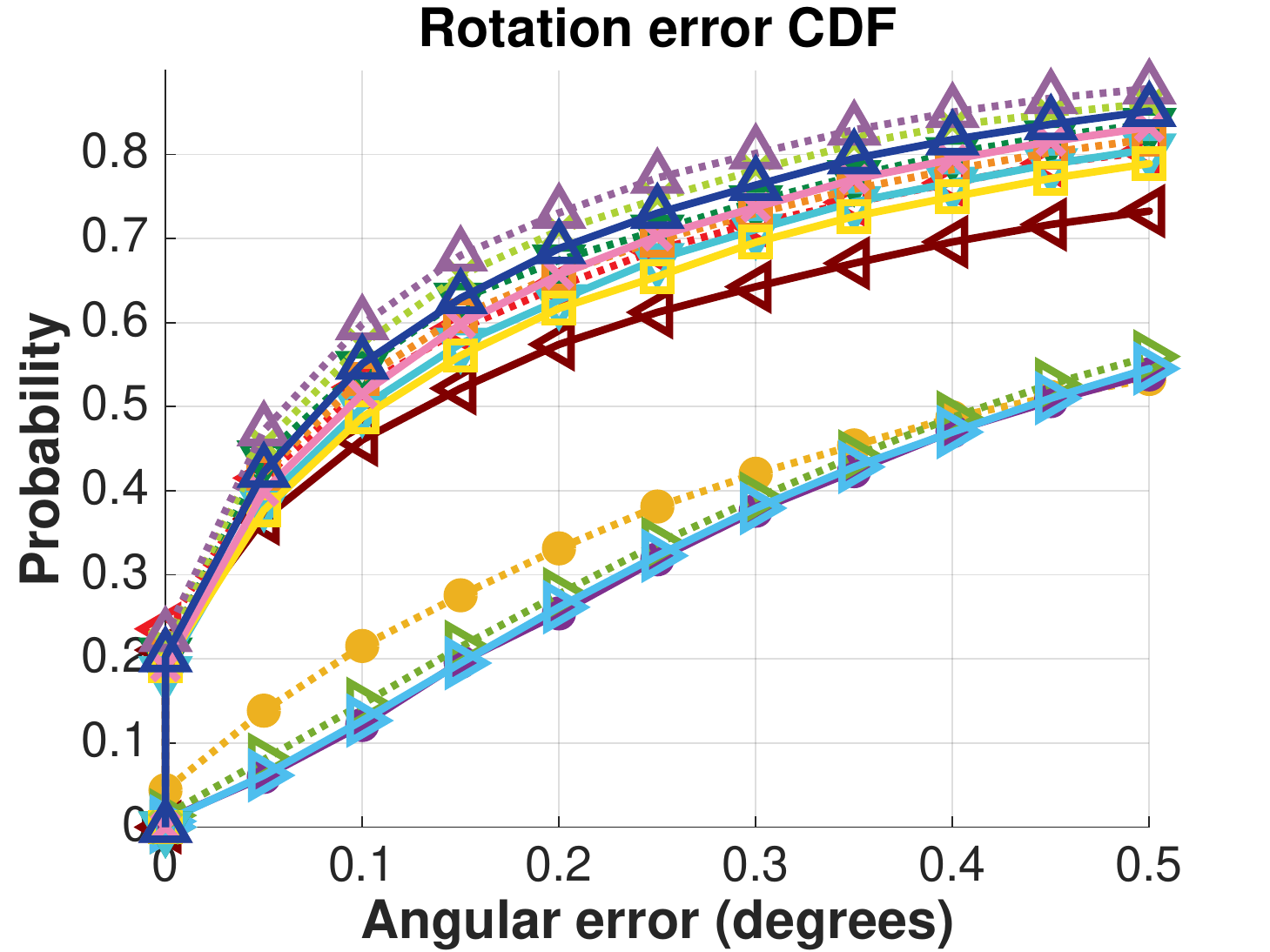}\\ 
	        \includegraphics[trim={1mm 0mm 10mm 0mm},clip, width=0.295\textwidth]{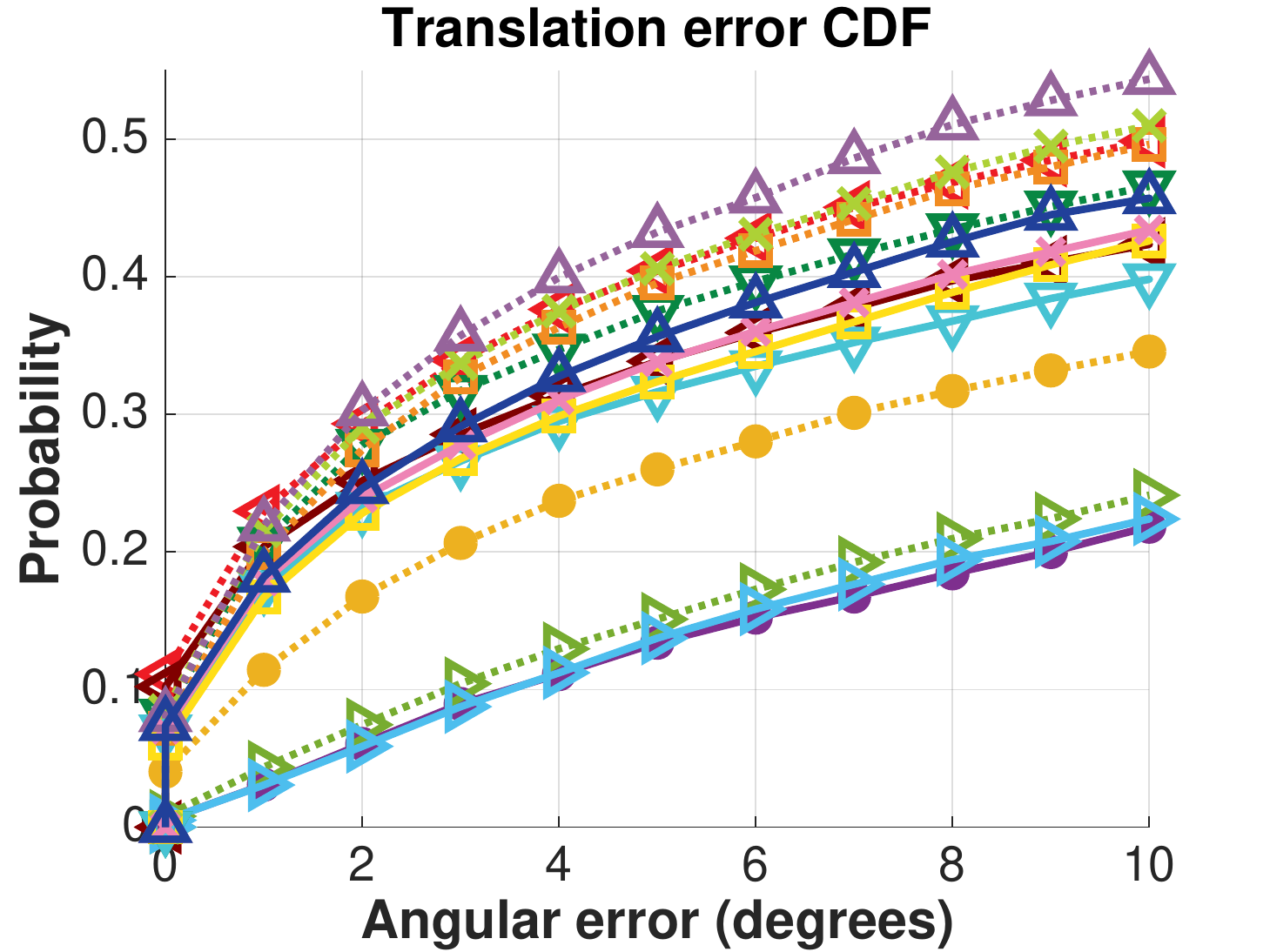}
	    \end{tabular}}
    \caption{ 
        The cumulative distribution functions of the rotation and translation errors (in degrees) of GC-RANSAC combined by different minimal and non-minimal solvers on datasets \Malaga ($9,064$ image pairs), \KITTI ($23,190$) and \ETH ($5,162$). 
        Being accurate is interpreted as a curve close to the top-left corner.}
    \label{fig:real_experiments}
\end{figure*}

The cumulative distribution functions (CDF) of the rotation and translation errors (in degrees) on the three tested datasets are shown in Fig.~\ref{fig:real_experiments}. 
Being accurate is interpreted as a curve close to the top-left corner.
The proposed OPT, LIN, OPT(S), and LIN(S) solvers are always among the top-performing methods. 
Interestingly, the LIN solver leads to the most accurate rotations and translations on these scenes. 
Since these datasets contain image pairs with relatively small baseline and, thus, small pose change between the frames, it is not a surprise that LIN is accurate.
Moreover, since the OPT solver finds the optimum of an algebraic error, which might not always coincide with the minimum of the geometric one, it can happen that LIN solver leads to better pose estimates than the OPT solver.
The average errors are reported in Table~\ref{table:real_experiments}.
All of the proposed solvers lead to more accurate results than the other compared methods.
The most accurate results are obtained by the combination of 5PC and LIN. 

\begin{table}[t]
\centering
\setlength{\tabcolsep}{4pt}\setlength\aboverulesep{0pt}\setlength\belowrulesep{0pt}%
\caption{ The avg.\ rotation and translation errors (in degrees) of GC-RANSAC~\cite{barath2017graph} combined with different pose solvers are reported on datasets {\fontfamily{cmtt}\selectfont Malaga}, {\fontfamily{cmtt}\selectfont KITTI}, {\fontfamily{cmtt}\selectfont ETH3D} (in total, $47,863$ image pairs). Columns 1--2 show the solvers used for minimal ($m$) and non-minimal ($>m$) fitting:
the proposed optimal solvers OPT/OPT(S), LIN/LIN(S), 3PC~\cite{ding_2020icra}, 5PC~\cite{stewenius2005minimal}, and 8PC~\cite{hartley1995defence}. The best results are shown in red, the second bests are in blue. The cumulative distribution functions of the errors are shown in   Fig.~\ref{fig:real_experiments}. }
\resizebox{1.00\columnwidth}{!}{ \small \begin{tabular}{ | c c || c c | c c | c c || c | }
 	\hline
 		& & \multicolumn{2}{c|}{\Malaga} & \multicolumn{2}{c|}{\KITTI} & \multicolumn{2}{c||}{\ETH} & \multirow{2}{*}{avg} \\
 		$m$ & $>m$ & $\epsilon_{\M R}$ & $\epsilon_{\M t}$ & $\epsilon_{\M R}$ & $\epsilon_{\M t}$ & $\epsilon_{\M R}$ & $\epsilon_{\M t}$ &  \\
 	\hline
    	\cellcolor{black!10}3PC & \cellcolor{black!10}OPT & \cellcolor{black!10}2.61 & \cellcolor{black!10}9.86 & \cellcolor{black!10}\win{0.04} & \cellcolor{black!10}{1.97} & \cellcolor{black!10}\second{0.33} & \cellcolor{black!10}\second{23.94} & \cellcolor{black!10}6.46 \\
    	\cellcolor{black!10}3PC & \cellcolor{black!10}OPT(S) & \cellcolor{black!10}3.02 & \cellcolor{black!10}10.17 & \cellcolor{black!10}\win{0.04} & \cellcolor{black!10}1.99 & \cellcolor{black!10}0.38 & \cellcolor{black!10}27.45 & \cellcolor{black!10}7.18 \\
    	\cellcolor{black!10}3PC & \cellcolor{black!10}LIN & \cellcolor{black!10}2.61 & \cellcolor{black!10}9.58 & \cellcolor{black!10}\win{0.04} & \cellcolor{black!10}{1.97} & \cellcolor{black!10}\win{0.31} & \cellcolor{black!10}\win{22.31} & \cellcolor{black!10}\second{6.14} \\
    	\cellcolor{black!10}3PC & \cellcolor{black!10}LIN(S) & \cellcolor{black!10}2.99 & \cellcolor{black!10}10.17 & \cellcolor{black!10}\win{0.04} & \cellcolor{black!10}1.96 & \cellcolor{black!10}0.40 & \cellcolor{black!10}25.59 & \cellcolor{black!10}6.86 \\
    	3PC & 3PC & 2.99 & 12.39 & 1.00 & 4.75 & 0.70 & 24.99 & 7.80 \\
    	3PC & 5PC & 2.89 & 6.77 & 0.25 & 3.22 & 1.82 & 34.60 & 8.26 \\
    	3PC & 8PC & \second{2.43} & 4.46 & 0.10 & 2.11 & 1.14 & 37.80 & 8.01 \\
    	\cellcolor{black!10}5PC & \cellcolor{black!10}OPT & \cellcolor{black!10}\win{2.41} & \cellcolor{black!10}3.59 & \cellcolor{black!10}\second{0.05} &  \cellcolor{black!10}1.90 & \cellcolor{black!10}0.41 & \cellcolor{black!10}28.71 & \cellcolor{black!10}6.18 \\
    	\cellcolor{black!10}5PC & \cellcolor{black!10}OPT(S) & \cellcolor{black!10}2.49 & \cellcolor{black!10}3.73 & \cellcolor{black!10}\second{0.05} & \cellcolor{black!10}1.91 & \cellcolor{black!10}0.44 & \cellcolor{black!10}31.11 & \cellcolor{black!10}6.62 \\
    	\cellcolor{black!10}5PC & \cellcolor{black!10}LIN & \cellcolor{black!10}\win{2.41} & \cellcolor{black!10}\second{3.49} & \cellcolor{black!10}\second{0.05} & \cellcolor{black!10}\second{1.89} & \cellcolor{black!10}0.37 & \cellcolor{black!10}26.86 & \cellcolor{black!10}\win{5.85} \\
    	\cellcolor{black!10}5PC & \cellcolor{black!10}LIN(S) & \cellcolor{black!10}2.47 & \cellcolor{black!10}3.74 & \cellcolor{black!10}\win{0.04} & \cellcolor{black!10}\win{1.88} & \cellcolor{black!10}0.46 & \cellcolor{black!10}29.32 & \cellcolor{black!10}6.32 \\
    	5PC & 3PC & 3.69 & 10.44 & 1.13 & 5.18 & 0.94 & 29.39 & 8.46 \\
    	5PC & 5PC & 2.72  & 4.79 & 0.27 & 3.30 & 2.15 & 36.74 & 8.33 \\
    	5PC & 8PC & 2.49 & \win{3.23} & 0.11 & 2.11 & 1.18 & 39.46 & 8.10 \\
 	\hline
\end{tabular} }
\label{table:real_experiments}
\end{table}

\subsection{Phone Dataset}

\begin{figure}[h]
		\includegraphics[width=0.325\columnwidth]{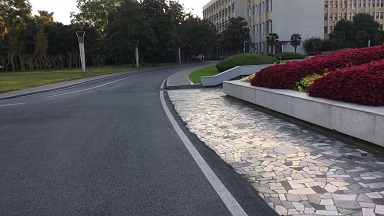}
		\includegraphics[width=0.325\columnwidth]{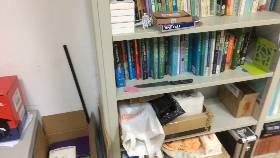}
		\includegraphics[width=0.325\columnwidth]{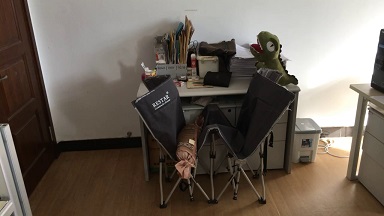}
	\caption{Sample images from the captured phone dataset. It consists a total of $10993$ image pairs with synchronized gravity directions, ground truth poses and 3D reconstructions. }
	\label{fig:example_phone}
\end{figure}

To further illustrate the usefulness of the proposed solvers, we tested them on images from a phone where the gravity directions are obtained from the built-in IMU.
Since we have not found such datasets, we decided to build one using images captured by a smartphone (iPhone 6s). We captured 6 sequences at @30Hz with the rear camera. The corresponding IMU data were captured at @100Hz with the phone's sensor (costs less than a dollar). We then synchronized the images and IMU data based on their timestamps. 
In order to obtain a ground truth, 
which can be used to test pose solvers, we applied the RealityCapture~\cite{realitycapture} software. 
A total of $10993$ image pairs with synchronized gravity direction, ground truth poses, calibrations and 3D reconstructions were obtained. Example images are shown in Fig.~\ref{fig:example_phone}.
We will make this dataset publicly available.

\begin{figure*}[t]
    \centering
    \includegraphics[width=0.8\textwidth]{figures/legend.pdf}\\
	\subfloat[]{
	\includegraphics[trim={1mm 0mm 10mm 0mm},clip, width=0.63\columnwidth]{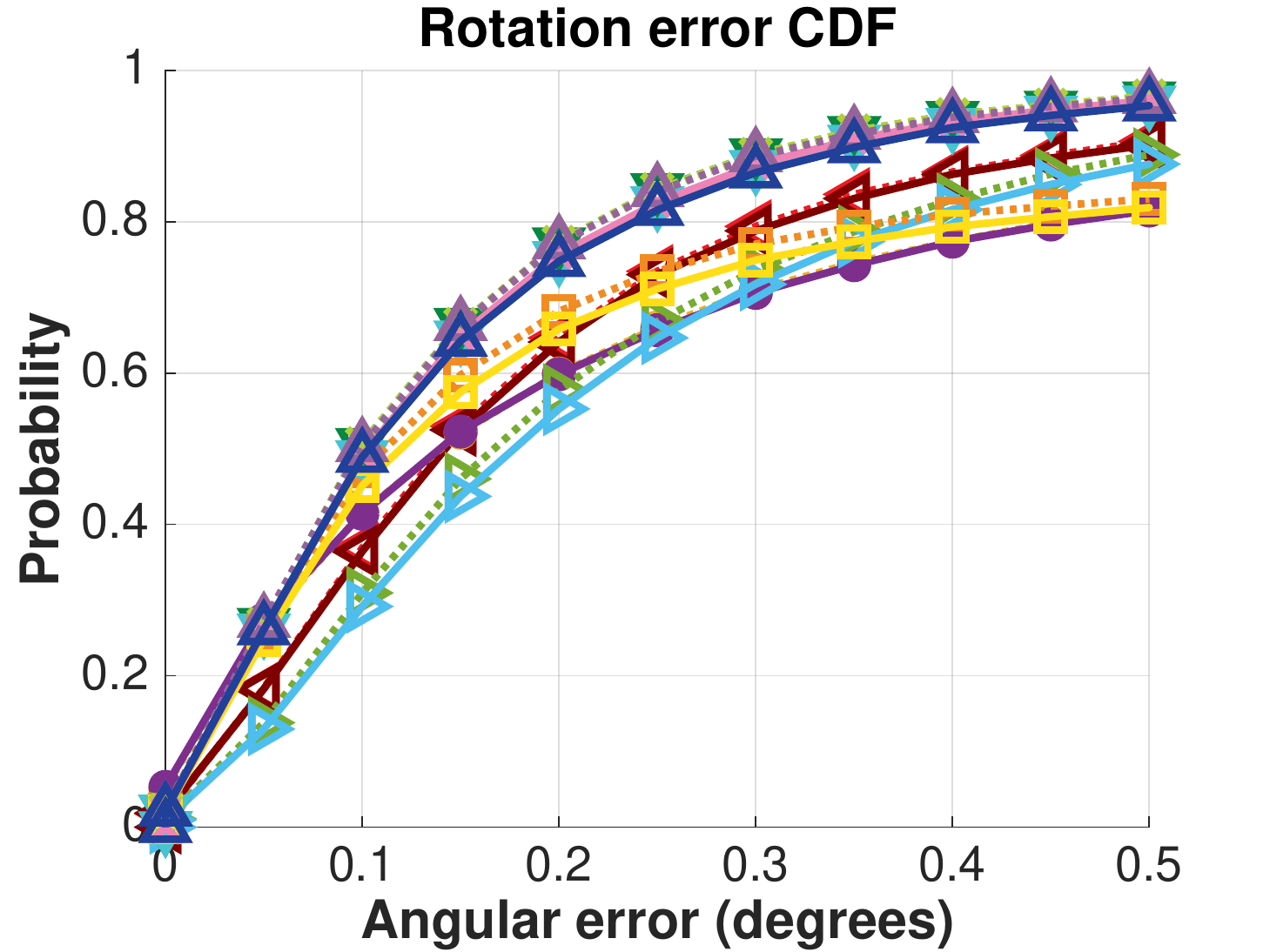}
	\includegraphics[trim={1mm 0mm 10mm 0mm},clip, width=0.63\columnwidth]{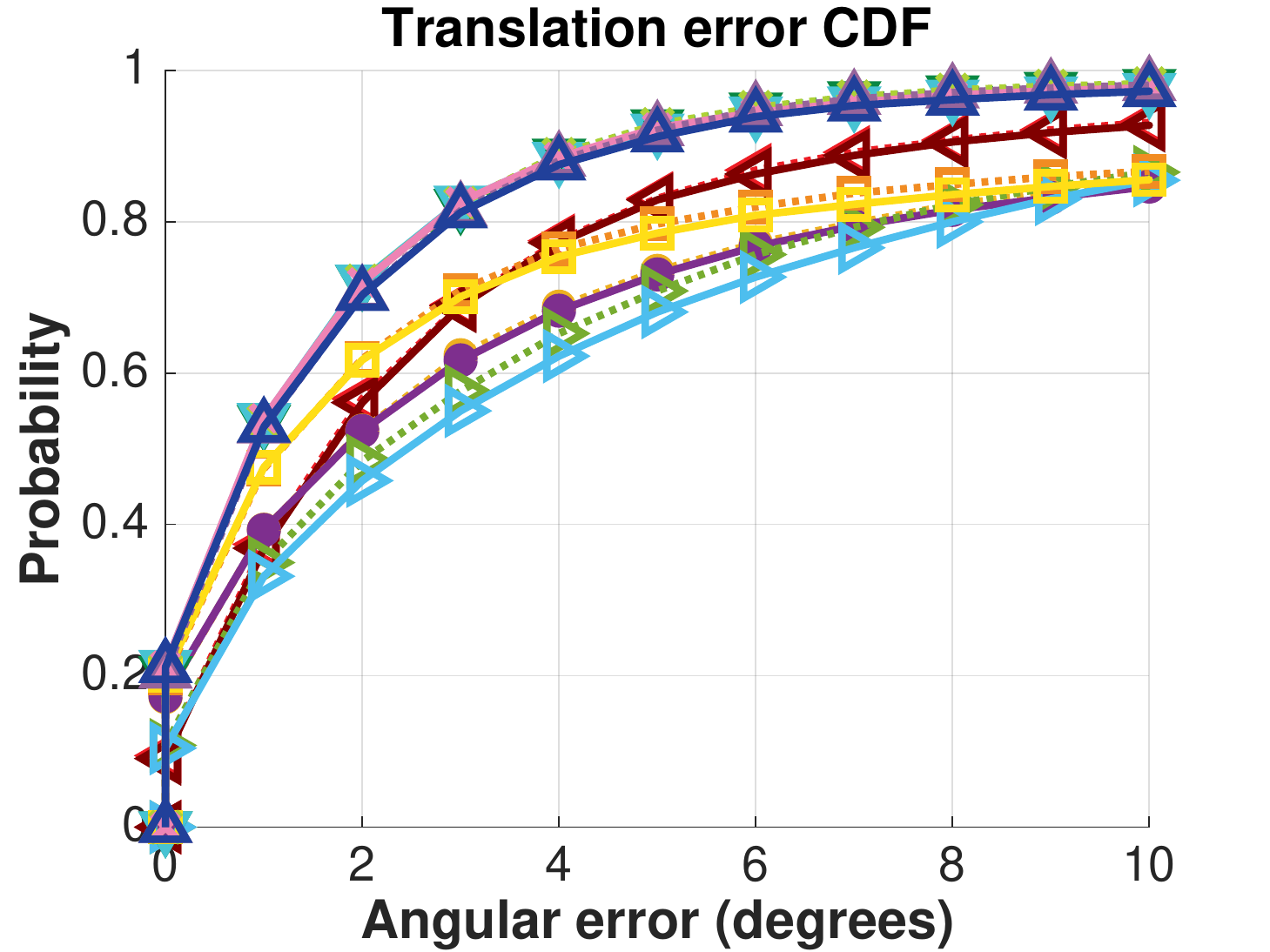}
    \label{fig:real_phone_experiments}
	}
	\subfloat[]{
	\includegraphics[trim={1mm 0mm 10mm 0mm},clip, width=0.63\columnwidth]{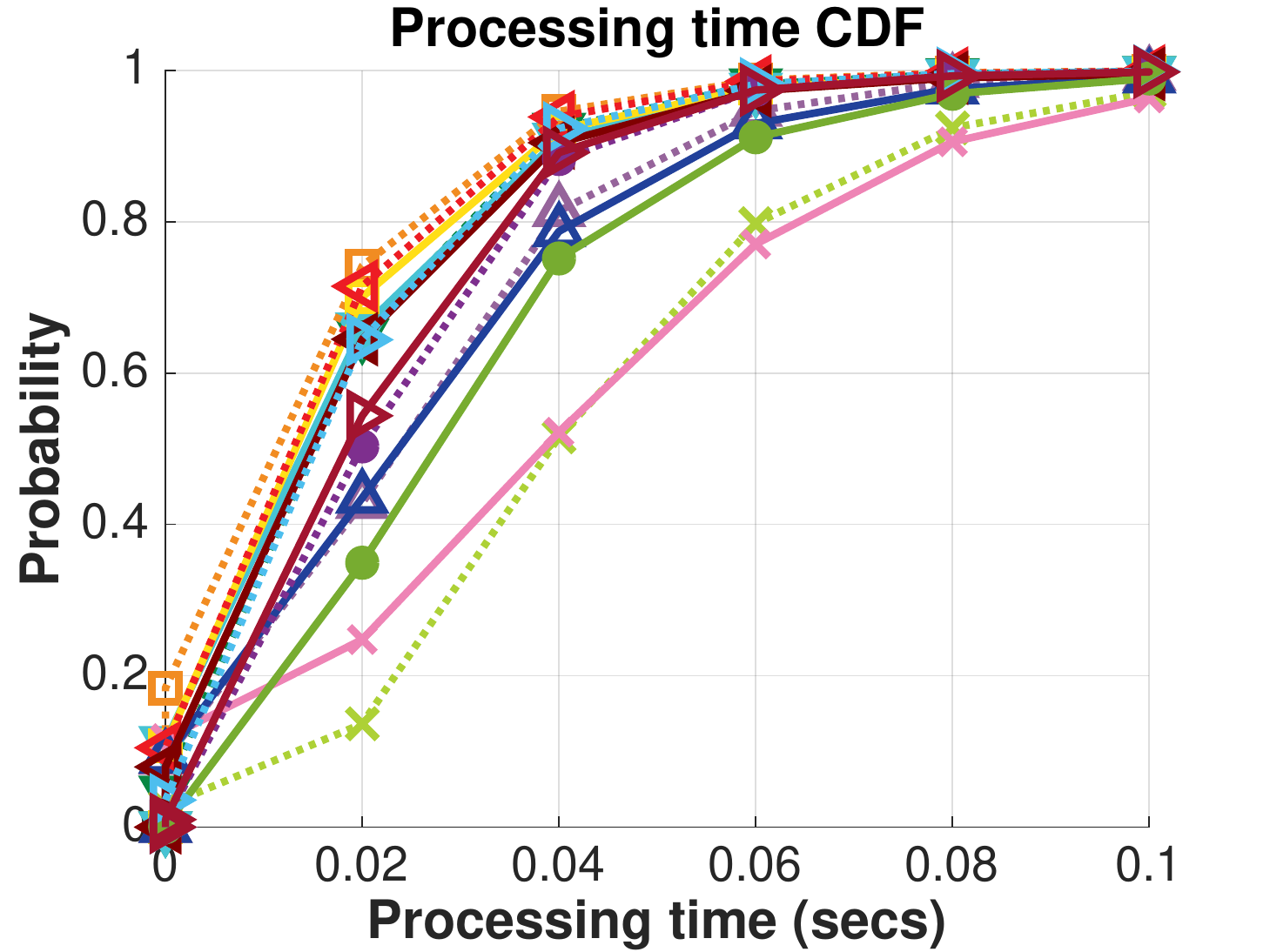}
    \label{fig:time_cdfs}
	}
    \caption{ 
        \textbf{(a)} The cumulative distribution functions (CDF) of the rotation and translation errors (in degrees) of GC-RANSAC combined by different minimal and non-minimal solvers on  the captured phone dataset  consisting of $10993$ image pairs. 
        Being accurate or fast is interpreted as a curve close to the top-left corner.
        \textbf{(b)} The CDF of the processing times (in seconds) of the robust estimation on all datasets ($47,863$ image pairs). }
\end{figure*}

The CDFs of the compared solvers are shown in the left two plots of Fig.~\ref{fig:real_phone_experiments}.
The same trend can be seen as before, {\ie}, the proposed optimal and linearized solvers lead to the most accurate rotations and translations. 
However, OPT and LIN leads to similar accuracy on this dataset. 
This is due to having slightly bigger view changes than in the datasets designed for autonomous driving. The bigger view changes increase the approximative nature of solver LIN and, thus, make its results marginally less accurate. Still, it is one of the top-performing methods in terms of accuracy.

\subsection{Processing Times}

The CDFs of the processing times of the robust estimation on all the $47863$ image pairs from all datasets are in Fig.~\ref{fig:time_cdfs}.
The proposed solvers based on Sturm sequences are the fastest ones. 
The optimal solver (OPT), compared to the other ones are the slowest method, while the linearized (LIN) solver has comparable speed to the other methods. 
It is important to mention that even the slowest combination  (\ie, 5PC+OPT) has an average run-time of $61$ ms.
Therefore, all tested combinations lead to \textit{real-time} performance.

\section{Extensions and Discussion}\label{ext}

\vspace{1mm}
\noindent\textbf{General case.}
The proposed technique can be, in theory, applied to the general case (without gravity prior). 
In this case, for the Cayley parameterization of the rotation, the elements of the matrix $\M C$ are polynomials in $x,y,z$. The three eigenvalues of $\M C$ have to satisfy~\eqref{q10}, but now there are three partial derivatives $\frac{\partial \alpha}{\partial x},\frac{\partial \alpha}{\partial y},\frac{\partial \alpha}{\partial z}$ that should be zero. This gives four polynomials with respect to four unknowns $\{\alpha,x,y,z\}$. Using Macaulay2~\cite{grayson2002macaulay}, we found that this system of polynomials has more than 2,000 solutions. 
A solver to such a system is impractical since the eigenvalue decomposition of a matrix of size $2,000\times2,000$ is extremely slow. 
However, it means that there are more than 2,000 stationary points.
Hence, locally optimized methods (\eg, \cite{Kneip2013DirectOO} or 8PC+LM) may  get stucked in these minima.

\noindent\textbf{Planar motion.}
In case of planar camera motion, which is usual scenario for unmanned ground vehicles with rigidly mounted cameras, we have constraint $t_y = 0$. 
Therefore, the middle row and column of the matrix $\M{C}$ in~\eqref{q9} can be removed. Matrix $\M{C}$ becomes a $2\times2$ matrix,
where the two eigenvalues should satisfy quadratic constraint
$\alpha^2-f_1\alpha+f_2=0$, where $f_1={{\operatorname {trace}}(\M C)}$ and $f_2= \det({\M C})$.
The remaining steps are similar to Sec.~\ref{poly}, and there are up to 8 solutions. In this case, the polynomial eigenvalue solution  needs to find the eigenvalues of a $10\times10$ matrix.

\noindent\textbf{Forward motion.}
In our experiments, we observed that the globally optimal solvers~\cite{zhao2020pami,Briales_2018_CVPR} estimate inaccurate poses in the case of forward motion. 
For this motion, the proposed globally optimal solvers return the best results. Due to the lack of space we included the experiments for forward motion in the supplementary material.

\section{Conclusions}

We propose globally optimal solutions for estimating the relative pose of two cameras, aligned by the gravity direction, from a larger-than-minimal set of point correspondences.
All of the proposed solvers, even the linearized ones, lead to results \textit{superior} to the state-of-the-art in terms of accuracy on approx.\ 50k image pairs from four widely-used datasets.
The techniques based on linearization or Sturm sequences lead to extremely fast robust estimation with an average of 22 ms run-time with no or not significant deterioration in the accuracy. 
Moreover, we captured a new dataset consisting of more than 10k photos taken by a smartphone with a gravity sensor. 

{\small
\bibliographystyle{ieee_fullname}
\bibliography{optimazition}
}

\end{document}